\pdfoutput=1
\documentclass{article}
\usepackage{caption}
\usepackage[lofdepth,lotdepth]{subfig}
\usepackage{arxiv}
\usepackage{graphicx}
\usepackage[utf8]{inputenc} 
\usepackage[T1]{fontenc}    
\usepackage[hide links]{hyperref}       
\usepackage{url}            
\usepackage{booktabs}       
\usepackage{amsfonts}       
\usepackage{amsmath}        
\usepackage{nicefrac}       
\usepackage{microtype}      
\usepackage{lipsum}
\usepackage[symbol]{footmisc}
\usepackage{pdfpages}
\usepackage{bm}

\makeatletter
\g@addto@macro{\UrlBreaks}{%
	\do\-}
\makeatother
\title{Wasserstein GAN: Deep Generation applied on financial time series}

\author{
	Pfenninger Moritz\textsuperscript{\textmd1} \\
	\href{mailto:pfennmor@students.zhaw.ch}{\texttt{pfennmor@students.zhaw.ch}} \\
	\And
	Bigler Nico\textsuperscript{\textmd1}\\
	\href{mailto:bigleda1@students.zhaw.ch}{\texttt{bigleda1@students.zhaw.ch}} \\
	\AND
	Samuel Rikli\textsuperscript{\textmd1}\\
	\href{mailto:riklisam@students.zhaw.ch}{\texttt{riklisam@students.zhaw.ch}} \\
	\And
	Joerg Osterrieder\textsuperscript{\textmd{1,2}}\\
	\href{mailto:joerg.osterrieder@zhaw.ch}{\textsuperscript{\textmd{1}}\texttt{joerg.osterrieder@zhaw.ch}}\\
	\href{mailto:joerg.osterrieder@utwente.nl}{\textsuperscript{\textmd{2}}\texttt{joerg.osterrieder@utwente.nl}}\\
	\AND
	\textnormal{\textsuperscript{\textmd{1}}School of Engineering}\\
	Zurich University of Applied Sciences\\
	Winterthur, Switzerland\\
	\AND
	\textnormal{\textsuperscript{\textmd{2}}The Hightech Business and Entrepreneurship Group}\\
	Faculty of Behavioural, Management and Social Sciences\\
	University of Twente\\
	Enschede, Netherlands\\
}

\begin{document}
	\maketitle
	\footnotetext[1]{\tiny{Financial support by the Swiss National Science Foundation within the project “Mathematics and Fintech - the next revolution in the digital transformation of the Finance industry” is gratefully acknowledged by the corresponding author. 
			This research has also received funding from the European Union's Horizon 2020 research and innovation program FIN-TECH: A Financial supervision and Technology compliance training program under the grant agreement No 825215 (Topic: ICT-35-2018, Type of action: CSA).
			Moreover, this article is also based upon the work from the Innosuisse Project 41084.1 IP-SBM Towards Explainable Artificial Intelligence and Machine Learning in Credit Risk Management.
			Furthermore, this article is based upon work from COST Action 19130 Fintech and Artificial Intelligence in Finance, supported by COST (European Cooperation in Science and Technology), www.cost.eu (Action Chair: Joerg Osterrieder).
			The authors are grateful to management committee members of the COST Action CA19130 Fintech and Artificial Intelligence in Finance as well as speakers and participants of the 5th European COST Conference on Artificial Intelligence in Finance and Industry, which took place at Zurich University of Applied Sciences, Switzerland, in September 2020.}}

	\begin{abstract}
		Modeling financial time series is challenging due to their high volatility and unexpected happenings on the market. Most financial models and algorithms trying to fill the lack of historical financial time series struggle to perform and are highly vulnerable to overfitting. As an alternative, we introduce in this paper a deep neural network called the WGAN-GP, a data-driven model that focuses on sample generation. The WGAN-GP consists of a generator and discriminator function which utilize an LSTM architecture. The WGAN-GP is supposed to learn the underlying structure of the input data, which in our case, is the Bitcoin. Bitcoin is unique in its behavior; the prices fluctuate what makes guessing the price trend hardly impossible. Through adversarial training, the WGAN-GP should learn the underlying structure of the bitcoin and generate very similar samples of the bitcoin distribution.
		The generated synthetic time series are visually indistinguishable from the real data. But the numerical results show that the generated data were close to the real data distribution but distinguishable. The model mainly shows a stable learning behavior. However, the model has space for optimization, which could be achieved by adjusting the hyperparameters.
		
		
	\end{abstract}

	\newpage
	\tableofcontents
	\newpage
	
	\section{Introductions}
	
	Sharing data has become more and more difficult due to stricter regulations around data management to control privacy requirements. Until recently, natural data is anonymized due to privacy risks. Nowadays, data can generated artificially using statistical parts of real data. Synthetic data generation has received considerable focus, owing to its benefits of addressing the data access restrictions by preserving the real data's multivariate relationships and statistical integrity. Synthetic data minimizes the disclosure risks and has the potential to show high usability by capturing the essential relationships and distributions from realistic data. Because of these reasons, the generation of fully synthetic data has gained more popularity and could be a next-step solution to real-world data sharing problems.
	
	Generative Adversarial Networks (GAN) are a group of unsupervised learning algorithms in computer science. Dr. Ian Goodfellow and his colleagues first introduced it in 2014 to build a generative adversarial Network architecture that creates new data instances that resemble the training data to create images.
	
	\begin{figure}[ht]
		\begin{centering}
			\includegraphics[width=4cm]{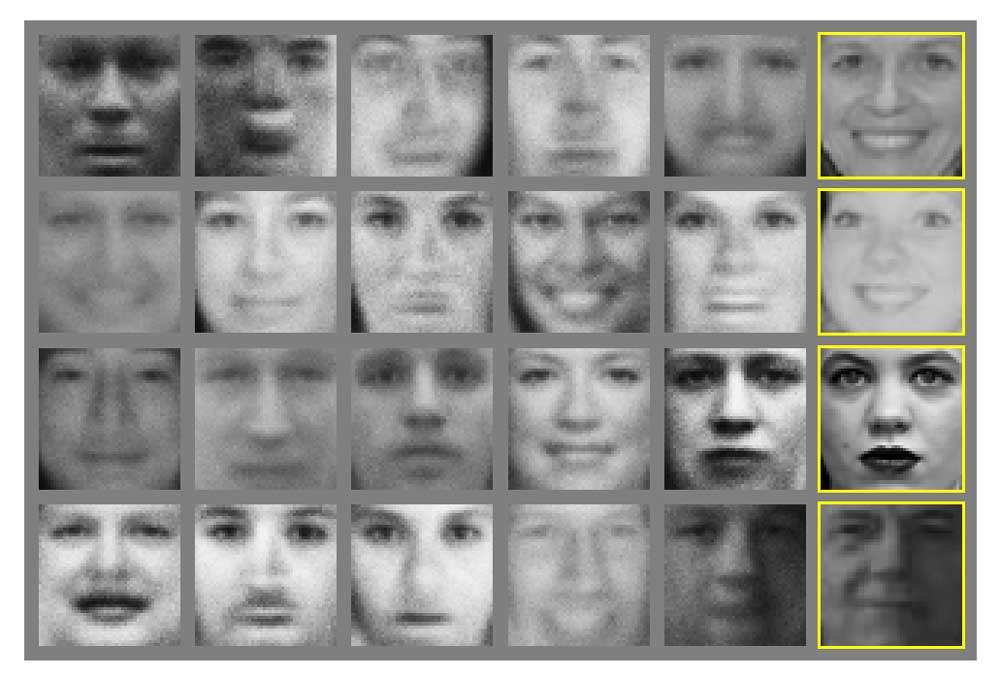}
			\caption{First-ever created fake images by a GAN [56]}
			\label{bild_gesicht}
		\end{centering}
	\end{figure}
	
	They can also use them to make data anonymous, which can be helpful to data containing personal content. There have already been approaches such as unknown data processing. The processing of personal data is a significant hurdle in machine learning. However, GANs are a promising approach for data set anonymization and can open the door for companies to process personal data. [9] While the first GANs focused on images with faces, the GANs are meanwhile able to generate ideas of almost any kind in high resolution. The first fake-generated images are demonstrated in [Fig. \ref{bild_gesicht}]. 
	It has been an important milestone in deep learning and the starting signal for further research on the topic of GAN. Over time, new implementations have been created that can testify to image quality and credibility in the authenticity of images. In the first place, it created low-resolution images, wherein the training process of the resolution is increased step by step. [10] [11]
	From all the GANs, there is only a fragment of the total GAN methods dealing with one-dimensional (1D) continuous-valued time series data. Throughout this paper, we reference to time series with a 1D continuous time-series database.
	
	In this paper, we like to carry out the following:
	
	\begin{itemize}
		\item{Introduction to the neural networks up to the GAN}
		\item{Implementation of the Wasserstein-GAN with gradient penalty and LSTM architecture applied on bitcoin.}
		\item{The results are evaluated by comparing the generated synthetic distribution against the historic data distribution.}
	\end{itemize}

	\newpage
	\section{Literature review}
	
\begin{figure}[htb]
		\centering
		\includegraphics[width=\textwidth]{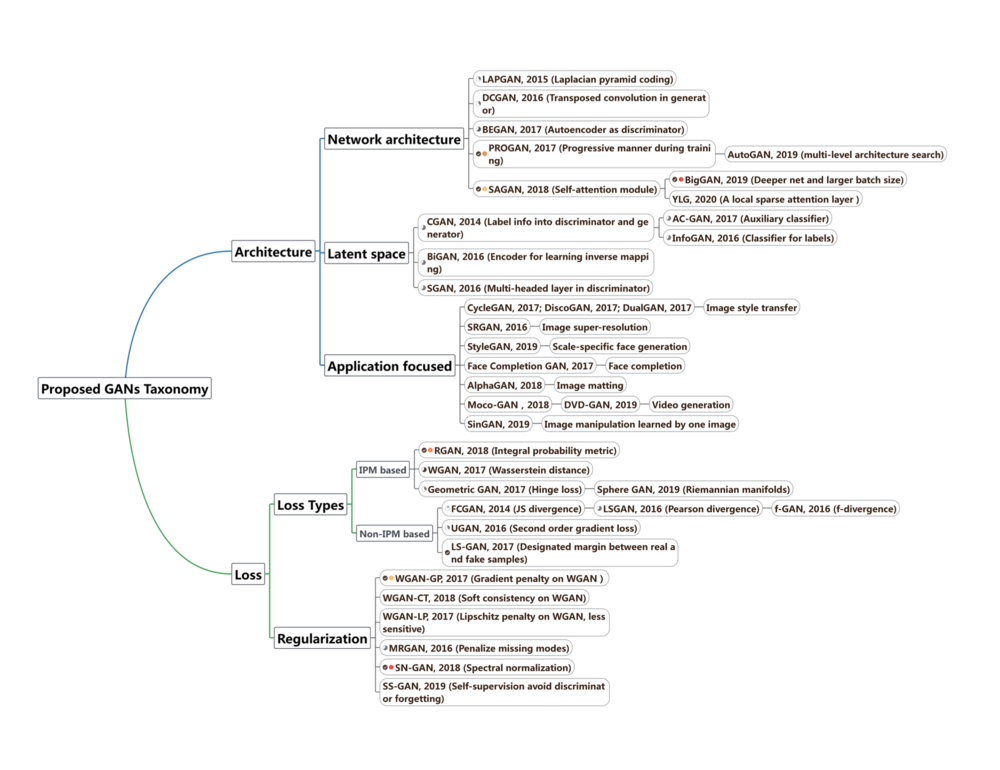}
		\caption{Different kind of GANs [57]}
		\label{fig:different kind of GANs}
	\end{figure}

	After a careful literature study on financial modeling, we understood that generating data may increase training models' performance in different situations across multidisciplinary financial activities [64]. To generate synthetic data, a model needs to learn the underlying distribution of the real data. The statistical properties of classic asset returns are a well-studied topic and referred to as Stylized Facts. The most important Stylized Facts can be captured in [59]. In this thesis, the goal is to synthesize Bitcoin's asset returns, which is the most traded cryptocurrency. It represents a high-risk investment and differs from traditional assets from a statistical point of view. Characterization of the parametric distributions of Bitcoin was provided by Stephen Chan et al. (2017) [60]. Time series [62] forecasting is a challenging task. There are many methods, such as the traditional ones like the ARIMA, AR, or the exponential smoothing, which only operate on a small-time series. Over the last decades, however, many companies have collected big sets of data [63]. The traditional methods that consider individual time series in isolation or operate on small-time series can fail to perform on such big data. This has brought the urge for new techniques to process all the available data and produce reliable forecasts. Deep neural networks have gained popularity in time series forecasting because they benefit from the massive amounts of data and may be a good alternative to many other machine learning and statistical techniques. [62] 
	The synthetization of Bitcoin returns will be accomplished by a rather new variant of GANs generative model. It is a class of machine learning frameworks designed by Ian Goodfellow. [56] There are already some papers that focus on the implementations of GAN to synthesize financial time series. Some of them are here shown. [66--68] There have been a lot more, of which some of them already achieved satisfactory results. In Figure \ref{fig:different kind of GANs}, there are all different kinds of GAN setups over time. Each implementation is somehow jointly responsible for the continuous further development of GANs.
	
	Because GANs often suffer from mode collapse during training, we introduce the improved GAN called Wasserstein GAN to improve learning stability. The papers [28--30] focus on implementing a Wasserstein GAN and show differences to the original GAN [57] and optimize them in specific ways. Although the WGAN makes progress toward stable training, it can sometimes fail to converge or generates poor sample quality. [28]
	
	There are a lot of neural networks, such as the feed-forward networks. Feed-forward neural networks associate input with output and allow the signal to travel only one way, where there are no feedback loops. For normal GAN architectures, it is difficult to detect any patterns in financial time series like Bitcoin. They change with time, and it is difficult to detect any patterns due to long-term trends, seasonal and cyclical fluctuations, and random noise. However, the biggest challenge is noise and the need to make them stationary. [61][65] Recurrent neural networks have signals that travel in both directions between the input and output layers, which makes them highly complicated as powerful. However, RNNs suffer from vanishing gradients in training. This problem can cause the model to slow down or even stall. Long-Short Term Memory networks (LSTM), created in 1990, can fix this problem due to their ability to store information in a more extended period and learn from inputs separated from each other by long time lags. [65] Long Short Term Memory Networks (LSTM) can capture patterns in the time series data, making them valuable tools to predict predictions regarding the future trend of the data.
	
	\section{Neural network}
	
\begin{figure}[ht]
		\centering
		\includegraphics[width=8.5cm]{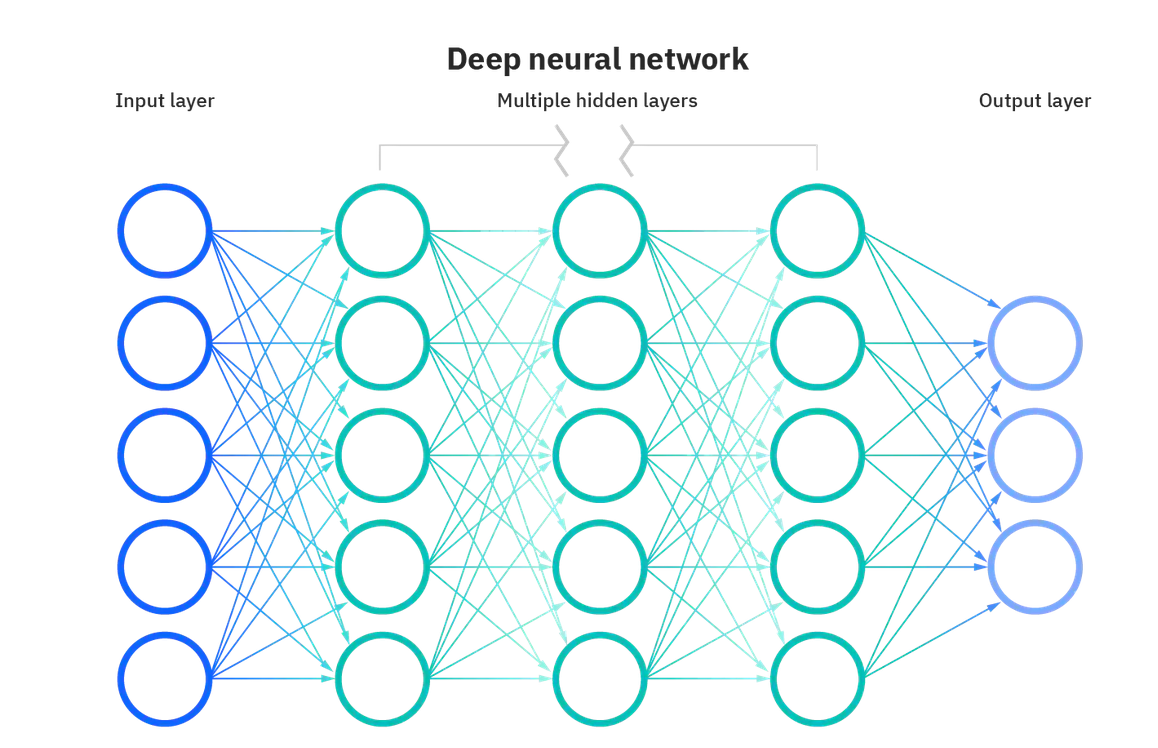}
		\caption{Structure of a deep neural network [13]}
		\label{fig:deep neural network}
	\end{figure}

	Neural networks are a subset of machine learning and are at the heart of deep learning algorithms. The human brain inspires their name and structure. Deep neural networks [Fig. \ref{fig:deep neural network}] compromises of node layers, containing an input layer, one or more hidden layers, and an output layer. Each node (also called perceptron) is connected to another and has an associated weight and threshold. If the output of any individual perceptron is above the specified threshold value, that perceptron is activated, sending data to the next layer of the network. Otherwise, no data passes along the next layer of the network.
	
	Neural networks rely on training data to learn and improve their accuracy over time. However, once these learning algorithms achieve high accuracy, they are potent tools in computer science and artificial intelligence, allowing us to classify and cluster data. [12]
	
	\subsection{Recurrent Neural Network}
	
	A recurrent neural network is a type of artificial neural network that uses sequential data or time-series data. It stores the output activations from the layers of the network. Then, the next time we feed an input example to the network, we include the previously-stored outputs as additional inputs. Like feed-forward, recurrent neural networks use training data to learn. Recurrent Neural Networks, however, can memorize previous inputs when a huge set of sequential data is fed. A deep neural network assumes that inputs and outputs are independent of each other. The output of a recurrent neural network depends on the primary elements within its sequences. [13][49]
	
	Even though RNNs are pretty powerful, they suffer from vanishing gradient problem, which hinders them from using long term information. They are useful for storing memories of 3 or 4 instances of past iterations. More enormous numbers of instances will not provide good results. [50]
	That is why we use a better variation of RNNs: Long Short Term Networks (LSTM).
	
	\begin{figure}[ht]
		\begin{centering}
			\includegraphics[width=11cm]{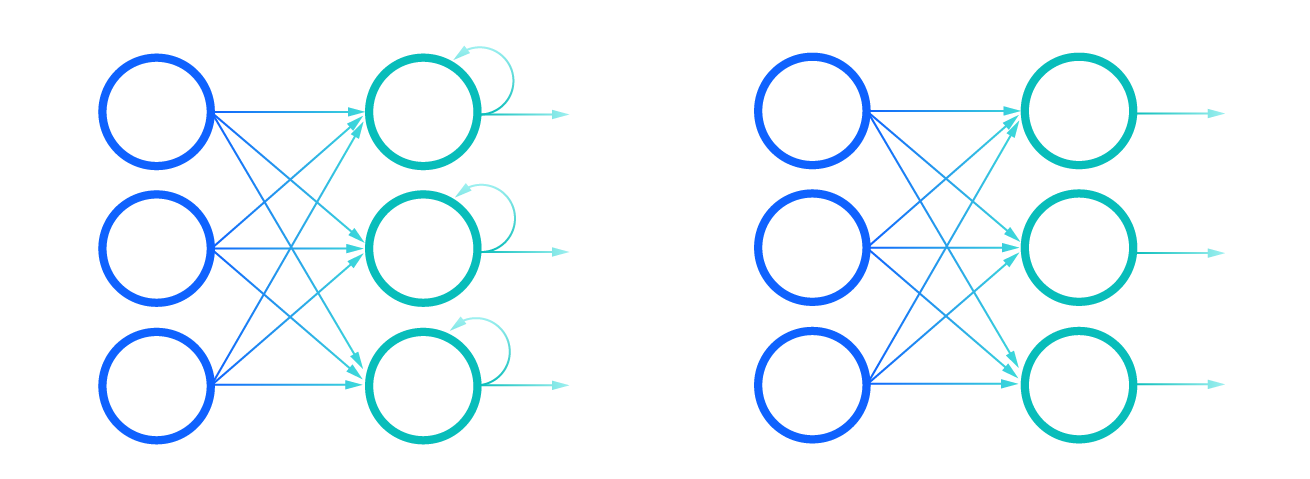}
			\caption{Recurrent neural network [13]}
			\label{Recurrent neural network}
		\end{centering}
	\end{figure}
	
	\subsection{Single Layer Perceptron (SLP)}
	
	In machine learning, the perceptron is an algorithm for supervised learning of binary classifiers, deciding whether an input vector is some specific class. Frank Rosenblatt firstly introduced the simplest form of a neural network in 1958. It consists of a single artifical neuron with adjustable weights and a treshold that will convert an input vector into an output vector and represent a simple associative memory. [13][14] The neuron consists of two mathematical functions: A calculation of the network input and an activation function that decides whether the calculated net input now "light up" or not. Therefore, it is binary in its output: Think of a small light that depends on the input values and weights, an input (sum) is formed, and then a function decides whether the light is lit. This concept of output generation is called feed-forward propagation. [51] Let's consider the structure of the perceptron. It contains four key components:
	
	\begin{figure}[ht]
		\begin{minipage}[m]{0.4\linewidth}
			\centering
			\includegraphics[width=\textwidth]{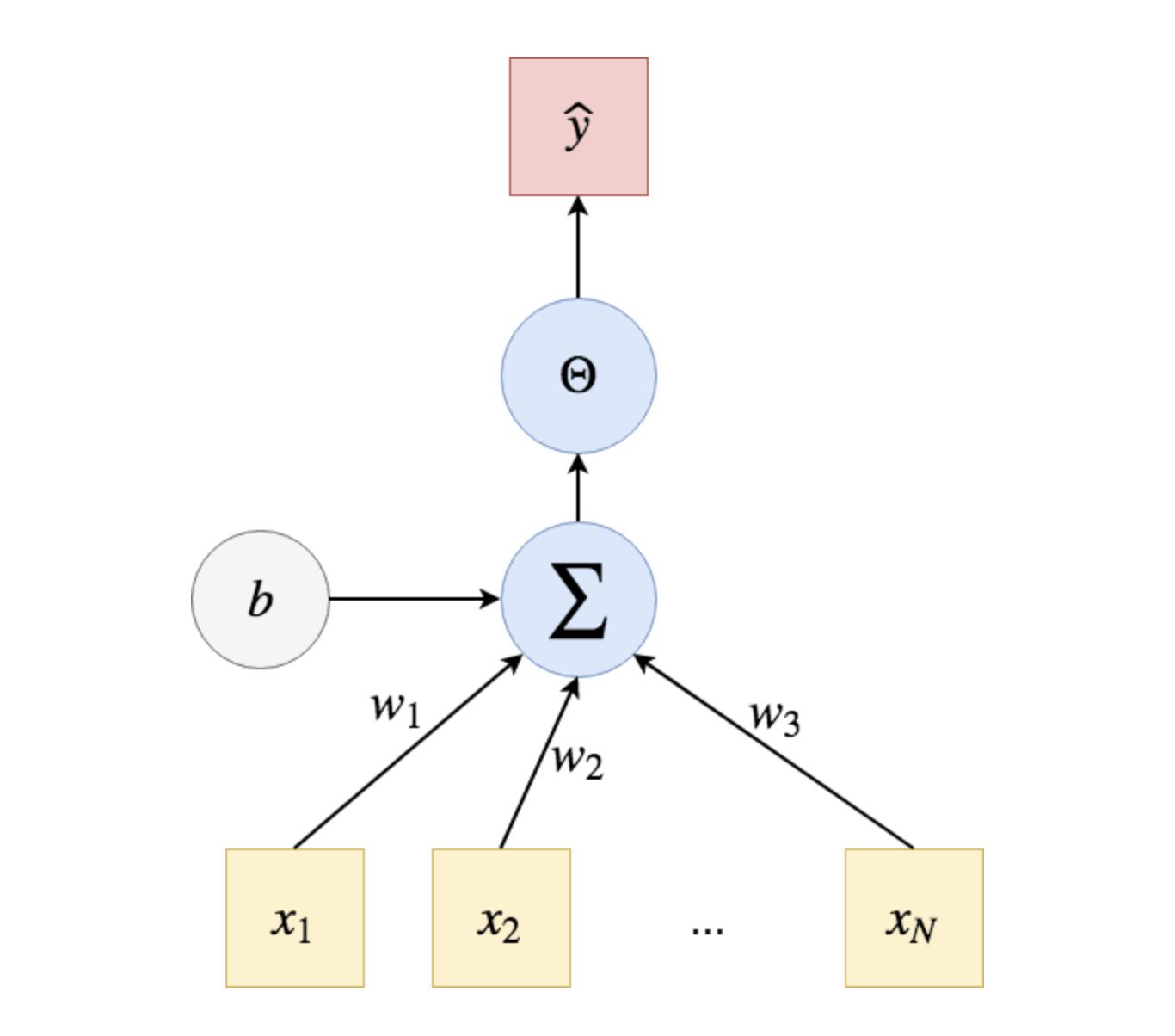}
			\caption{Computational graph}
			\label{fig:perceptron}
		\end{minipage}
		\hspace{0.5cm}
		\begin{minipage}[b]{0.5\linewidth}
			\centering
			\begin{enumerate}
				\item{input {$x_i$}}
				\item{weights {$w_i$}}
				\item{weighted sum $\sum$}
				\item{thresholding value by using an activation function}
			\end{enumerate}
		\end{minipage}
	\end{figure}
		
	By using the weighted summing function, the perceptron becomes a learnable parameter. By adjusting the weights, the perceptron can differentiate between two classes and therefore model the classes. The weighted sum is transmitted through an activation function $\theta$. Activation functions decide to activate or deactivate the neurons to get a desired output. 
	The $\sum$ represents the linear combination of the inputs $X(x_1,x_2)$ of length $n$ that is weighted by a weight vector $w$.
	\begin{equation}
	x\textsubscript{1}  = x\textsubscript{1} * w\textsubscript{1}
	\end{equation} 
	\begin{equation}
	x\textsubscript{2}  = x\textsubscript{2} * w\textsubscript{2}
	\end{equation}
	The output of the activation functions is the output of the perceptron. The larger the numerical value of the output will be, the higher the confidence of the prediction. [16--19]
	Let us simplify. The inputs are defined as $x_1$ and $x_2$. Each input $x_1$, $x_2$ gets multiplied by its assigned weight $w_1$ and $w_2$.
	The input vector represents the training data on which the neural network is getting trained. Once this vector is defined, each element $x\textsubscript{i}$ of the vector is assigned to a weight $w\textsubscript{i}$. The input elements are multiplied by their associated weight, and the resulting $\textsl{n}$ results are summed to obtain a weighted average. The significance of the impact on individual input vector elements on the weighted average depends on the individual weights. After the weighted average is calculated, a bias is added. The bias is an element that adjusts the boundary away from origin without any dependence on the input value:
	\begin{equation}
	v = (x\textsubscript{1} * w\textsubscript{1}) + (x\textsubscript{2} * w\textsubscript{2}) + b
	\end{equation}
	Then, perceptron's output will be presented by transforming it to a (non-)linear activation function. We explain later what the activation functions are crucial. [58]
	\begin{equation}
	{y} = {f}(x\textsubscript{1} * w\textsubscript{1} + x\textsubscript{2} * w\textsubscript{2} + b)
	\end{equation}
	The SLP was, in a way, the beginning of artificial intelligence and mostly inspired the development of more complex neural networks. The main limitation of the SLP models is that those perceptron models are only accurate when working with data that is linearly separable and is therefore limited. [18]
	
	\subsection{Multi layer perceptron}
	
\begin{figure}[ht]
		\begin{centering}
			\includegraphics[width=15cm]{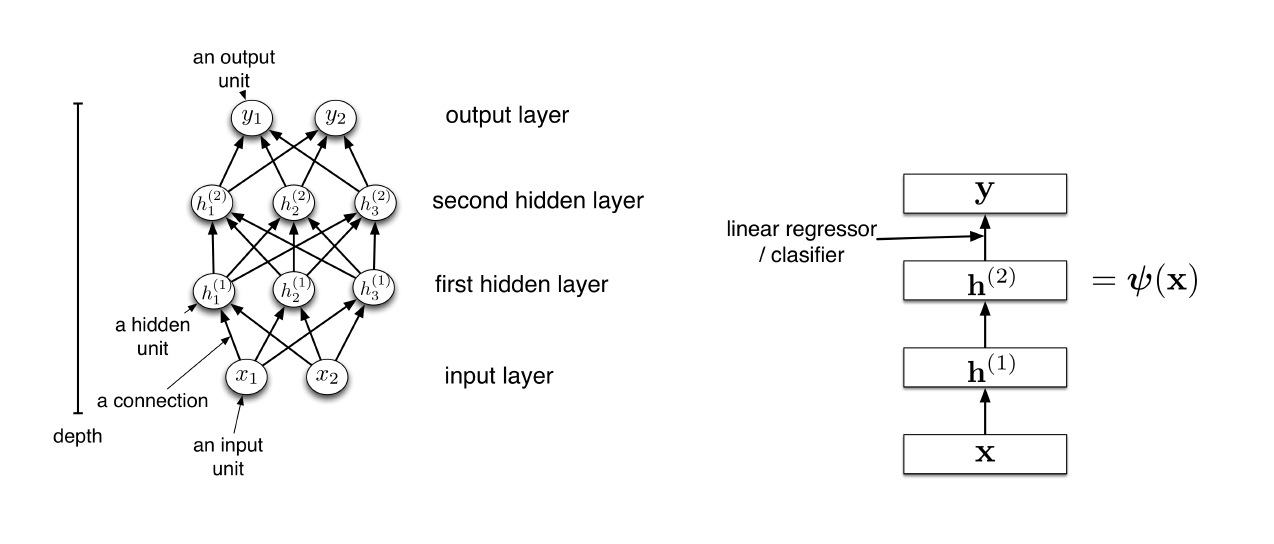}
			\caption{A multilayer perceptron with two hidden layers}
			\label{fig:multilayer perceptron}
		\end{centering}
	\end{figure}

	MLPs are distinguished from SLPs because they are hidden layers that affect the output of the model. In the MLP, the neurons are structured into layers. They belong to the class of feed-forward neural networks (FNNs), which means that nothing ever flows back during output generation, i.e. everything flows from input to output. In Figure \ref{fig:multilayer perceptron}, we can see an artificial neural network with multi-layer perceptrons. The multi-layer perceptrons contain inputs $x_n$, hidden layers $h_n\textsuperscript{(n)}$, and output layers $y_n$.
	
	Each neuron in the hidden layer $h_n\textsuperscript{(n)}$ is connected to each other neuron in $u_{n+1}\textsuperscript{(n+1)}$. The hidden layers consist of many neurons with connections between the layers. Each input value is associated with a weight connected to each in the hidden layers. The number of layers is known as the depth, and the number of units in a layer is known as the width. The last layer of neurons holds the individual classification or regression output. [6] \newline 
	The general neuron processing unit looks like this:
	
	\begin{equation}
	a = \phi(\sum\textsubscript{j} w\textsubscript{j} * x\textsubscript{j} + b)
	\end{equation}
	
	Where $x_j$ are the inputs to the unit, the $w_j$ are the weights, b is the bias, a is the unit's activation, and $\phi$ the non-linear activation function. The possibilities of machine learning are immense. Here are just some examples:
	
	\begin{itemize}
		\item{Regression uses a linear model, so $\phi$(z) = z.}
		\item{In binary linear classifiers, $\phi$ is a hard threshold at zero.}
		\item{In logistic regression, $\phi$ is the logistic function $\phi$ (z) = 1 = (1 + e + \textsuperscript{-z})}
	\end{itemize}
	
	\newpage
	A neural network is a combination of several of these presented units. Each one takes on an effortless action. In aggregation, it is capable of powerful and useful computations.
	As mentioned before, every unit in one layer is connected to every unit in the next layer. Each unit has its own bias and weight for every unit's connections in the two consecutive layers. We can define the first hidden layers $h_\textsubscript{i}$\textsuperscript{(n)} as follows:
	
	\begin{equation}
	\centering
	h\textsubscript{i}\textsuperscript{(1)} = \phi\textsuperscript{(1)}\Bigr(\sum_{j} w\textsubscript{i}\textsubscript{j}\textsuperscript{(1)}x\textsubscript{j} + b\textsubscript{i}\textsuperscript{(1)}\Bigr)
	\end{equation}
	
	\begin{equation}
	h_i\textsuperscript{(2)} = \phi\textsuperscript{(2)}\Bigr(\sum_{j} w\textsubscript{i}\textsubscript{j}\textsuperscript{(2)}h\textsubscript{j}\textsuperscript{2} + b\textsubscript{i}\textsuperscript{(1)}\Bigr)
	\end{equation}
	
	\begin{equation}
	h\textsubscript{i}\textsuperscript{(3)} = \phi\textsuperscript{(3)}\Bigr(\sum_{j} w\textsubscript{i}\textsubscript{j}\textsuperscript{(3)}h\textsubscript{j}\textsuperscript{2} + b\textsubscript{i}\textsuperscript{(3)}\Bigr)
	\end{equation}
	
	Consider distinguishing $\phi(1)$ and $\phi(2)$ because the different layers may have other activation functions. \newline Each layer contains multiple units, which represent the activations of all its units with an activation vector. There is a weight for every pair of units in two consecutive layers. We represent each layer's weight for every pair of units in two consecutive layers with a weight matrix. So the above computations could now be written in the desired vectorized form:
	
	\begin{equation}
	\centering
	h\textsuperscript{(1)} = \phi\textsuperscript{(1)}\Bigr(\textbf{W}\textsuperscript{(1)}\textbf{x} + \textbf{b}\textsuperscript{(1)}\Bigr)
	\end{equation}
	
	\begin{equation}
	h\textsuperscript{(2)} = \phi\textsuperscript{(2)}\Bigr(\textbf{W}\textsuperscript{(2)}\textbf{h}\textsuperscript{(1)} + \textbf{b}\textsuperscript{(2)}\Bigr)
	\end{equation}
	
	\begin{equation}
	y_i = \phi\textsuperscript{(3)}\Bigr(\textbf{W}\textsuperscript{(3)}\textbf{h}\textsuperscript{(2)} + \textbf{b}\textsuperscript{(3)}\Bigr)
	\end{equation}
	
	By applying the activation function to a vector, it independently happens to all the entries. In the feed-forward neural networks, the units are arranged into a graph without cycles, so all the computation can be done sequentially. However, in a recurrent neural network, the graph contains cycles, so the processing can feed into itself and is therefore suitable for time series. [7]
	
	\subsection{Backpropagation}
	
	Backpropagation is an algorithm of supervised learning of artificial neural networks using gradient descent. With an error function, the method calculates the gradient of the error function concerning the neural network's weights.
	The calculation runs backward through the whole network, with the gradient of the final layer of weights being calculated first and then the gradient of the first layer of weights. Partial computations of the gradient from one layer are reused to compute the gradient for the previous layer. [47]
	
	\subsection{Activation functions}
	
\begin{figure}[ht]
		\begin{centering}
			\includegraphics[width=7cm]{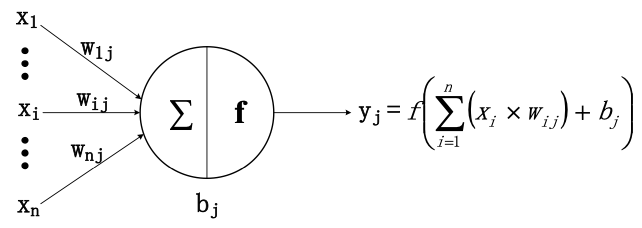}
			\caption{Structure of an activation function}
			\label{fig:activation function}
		\end{centering}
	\end{figure}

	An activation function is a unit that determines which information should be transmitted to the next neuron. Each neuron in the neural network accepts the output value of the neurons from previous layers as input and passes the processed value to the next layer. In a multi-layer neural network, there is a function between these two layers. This function is called the activation function. [Fig. \ref{fig:activation function}] There are several different activation functions, which can output different outputs within a range.
	
	Using a linear function, the input of each layer will be a linear function of the output to the previous layer. The network would be limited in learning linear functions, thus modeling them. No matter how many the neural network has, the output would always be a linear combination. They determine the accuracy of a deep learning model and also the computational efficiency of training a model. Many activation functions are employed, of which ReLu, sigmoid and tanh are extensively used. [8] The $\tanh$ function outputs a natural number between [-1,1]. Because the mean value of the output is 0, it achieves a normalization to make the following layers easier to learn. The sigmoid function is a non-linear activation function used for binary classification problems by transforming the output between [0, 1]. By $x$ approaching $\theta$, the gradient becomes steeper.
	In ReLu, when $x$ is less than 0, its function value is 0. When $x$ is greater than or equal to 0, its function value is $x$ itself. When $x$ is less than 0, the gradient of ReLU is 0, which means the back-propagated error will be multiplied by 0. [8]
	
	\subsection{Optimizers - RMSprop}
	
	Optimizers are algorithms that change the attributes of weights and learning rates to minimize their losses. [23] The way of changing the weights or the learning rate is defined by the optimizer you are using. A good example helps to imagine how an optimizer is working. A mountain climber wants to climb down from a mountain but cannot see anything. She does not know where to walk, but she begins to feel her way. By taking a step, she thinks whether it is going down or not. With every step upwards, she makes a loss of progress. With every step downwards, she makes progress, because she wants to reach the bottom. If she does this over some time and only carries out the downhill steps, she will eventually get her destination at the bottom. This process is also done by optimizing a loss function. Optimization algorithms are reducing the losses and provide the most accurate results possible. 
	
	\subsubsection*{RMSprop}
	
\begin{figure}[ht]
		\centering
		\includegraphics[width=7.8cm]{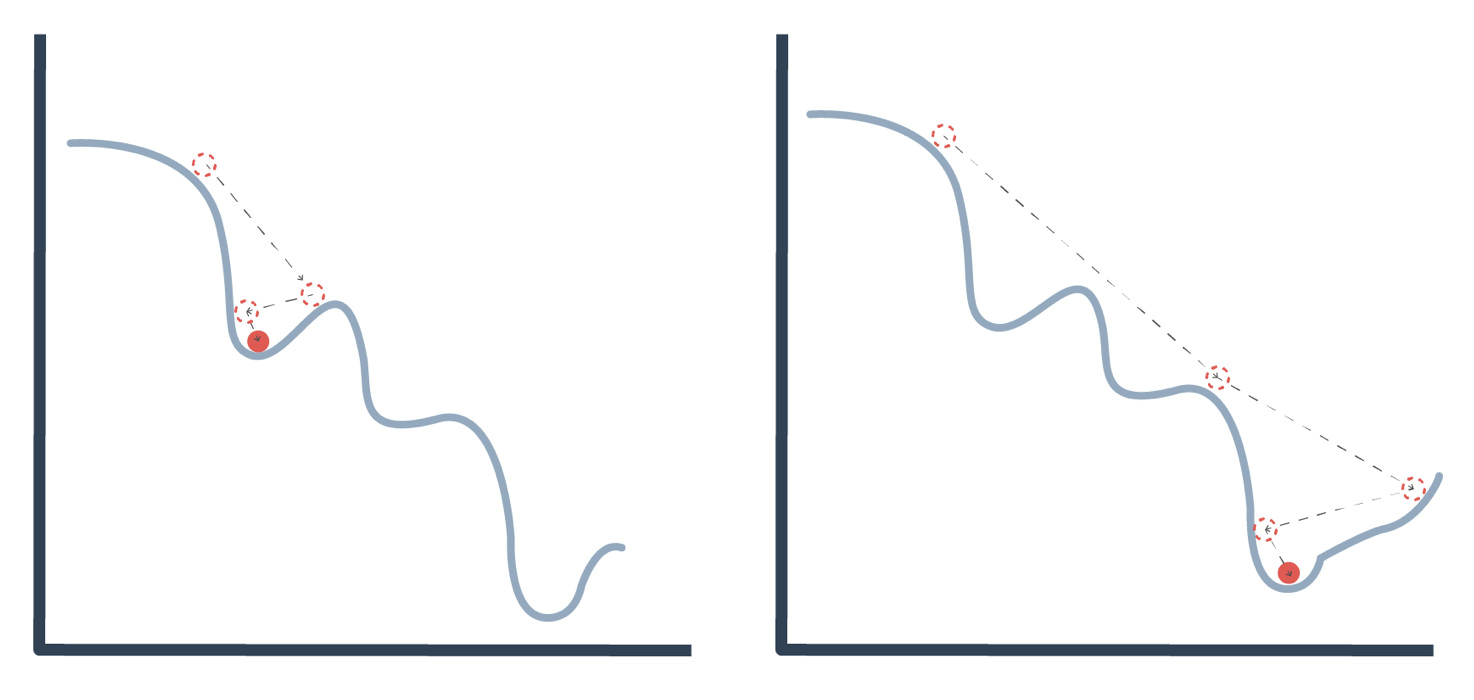}
		\caption{Learning rate}
		\label{fig:learning rate}
	\end{figure}

	RMSprop divides the learning rate by an exponentially decaying average of squared gradients. It decides how much of the gradient you update.
	More immense steps mean that the weights are changed more every iteration to reach their optimal value faster but possibly miss the exact optimum. So with smaller steps, the weights are changed less every iteration, so it takes more epochs to reach their optimal value, but they are less likely to miss the optimum loss function's optimum. In Figure \ref{fig:learning rate}, we can see that if the red dot rolls carefully with a small learning rate, we can expect consistent progress. [24]
	
	There is an option of learning rates which allows, to use of bigger steps in the first epochs and then reduce the step size when the weights have come in proximity of their optimal value.
	Several optimizers were introduced in the last few years, where every optimizer has its advantages and disadvantages. The players' weights are updated according to a stochastic gradient descent step with either ADAM or RMSprop optimizer to increase training convergence. It is usually used as a learning criterion of the optimization problem. [24] Remember the RMSprop optimizer, which we will later use in the Implementation. ADAM is one of the most used optimizers, but it did not give better results, which is why we do not go further into the ADAM optimizer.
	
	\subsection{Loss functions}
	
	Deep learning neural networks use a stochastic gradient descent optimization algorithm. The error of the actual state must be estimated repeatedly. That case requires a loss function to estimate the loss to update the weights, and reduce the loss on the following evaluation. The loss function takes a batch of actual samples and generated samples and then calculates the difference. The lower our loss is, the better the performance of our model. They are explained later in more detail. [27]
	
	\subsection{Learning rate}
	
	To produce stabilized GAN models, the learning rates should be low. The movement is across the gradient slope by taking smaller values, so the local minimum does not get missed. But higher learning rates might cause the gradient descent to overshoot the minimum, and as a result, the resulting model fails to converge to a minimum and leads to training failure of GANs. [55]
	
	\subsection{Batch size}
	
	The batch size determines the number of instances passed through the model, before the backpropagation for each epoch happens. A Smaller batch size results in updating the error gradients based on a smaller batch of samples and offers a regularization effect to reduce the error. Sometimes, increasing the minibatch size can improve the performance of the model. However, bigger batch size is expected to impact the performance because of the training of the Discriminator. By using a lot of samples, will end in overpowering the Generator. [55]
	
	\subsection{Long Short-Term Memory}
	
	The recurrent neural network suffers from gradient vanishing and exploding problems. But this is where the Long Short-Term Memory (LSTM) comes in. The LSTM networks are a modified version of recurrent neural networks (RNN making it easier to remember past data in memory. The vanishing gradient problem of RNN is resolved here. LSTM is well-suited to classify, process, and predict time series. It trains the model by using backpropagation. In a LSTM network, three gates are present, as in Figure \ref{fig:LSTM}:
	
	\begin{figure}[ht]
		\begin{centering}
			\includegraphics[width=7cm]{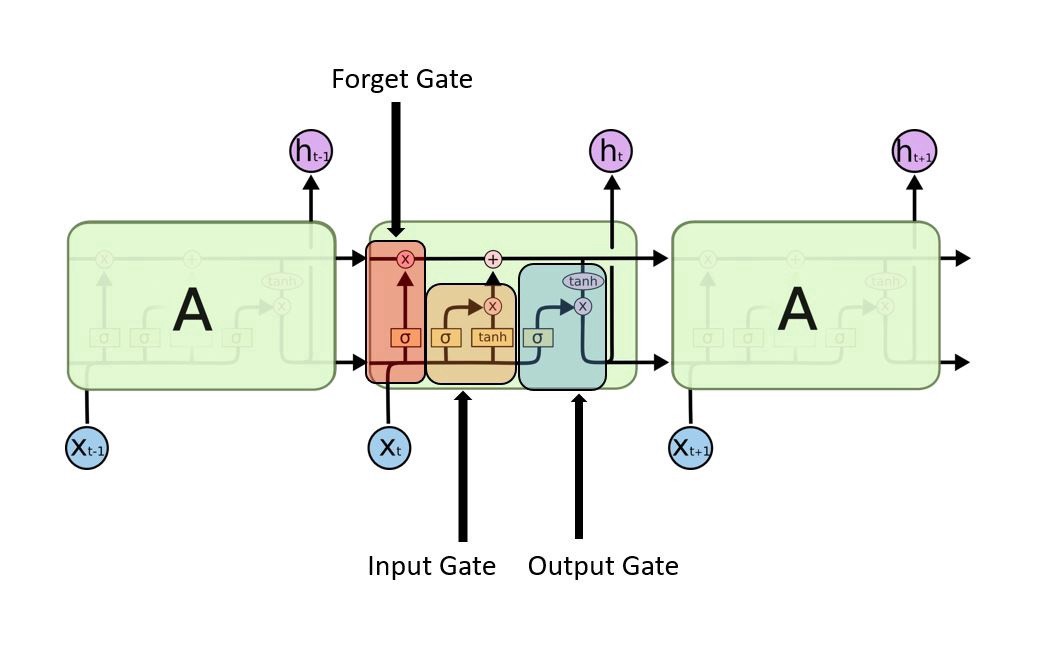}
			\caption{Structure of a LSTM}
			\label{fig:LSTM}
		\end{centering}
	\end{figure}
	
	\textbf{Input}
	\newline
	The Input gate discovers which value from input should be used to modify the memory. The sigmoid function then decides which values get through [0,1]. The tanh function gives weightage to the values which are passed, deciding their level of importance ranging from [-1, 1].
	\begin{equation}
	i\textsubscript{t} = \sigma(W\textsubscript{i} * [h\textsubscript{t-1}, x\textsubscript{t}] + b\textsubscript{i}
	\end{equation}
	\begin{equation}
	\tilde{C}\textsubscript{t} = tanh (W\textsubscript{c} * [h\textsubscript{t-1}, x\textsubscript{t}] + b\textsubscript{c})
	\end{equation}
	\textbf{Forget gate} 
	\newline
	The forget gate shows what details are discarded from the block. It is decided by the activation function. It considers the previous state h$\textsubscript{(t-1)}$ and the content input ($X\textsubscript{t}$) and outputs a number between [0,1] for each number in the cell state C$\textsubscript{(t-1)}$.
	
	\begin{equation}
	f\textsubscript{t} = \sigma(W\textsubscript{f} * [h \textsubscript{t-1}, x\textsubscript{t}] + b\textsubscript{f})
	\end{equation}
	
	\textbf{Output gate} 
	\newline 
	The input and the memory of the block is used to decide the output. Activation functions decide which values to let through [0, 1], and then the tanh function gives weightage age to all values, which then decides its value from [-1, 1] and multiplies with the output of the activation function. [49]
	
	\begin{equation}
	o_t = \sigma(W\textsubscript{o}[h\textsubscript{t-1}, x\textsubscript{t}] + B\textsubscript{o})
	\end{equation}
	
	\newpage
	\section{Generative Adversarial Network (GAN)}
	
	\begin{figure}[ht]
		\begin{centering}
			\includegraphics[width=13cm]{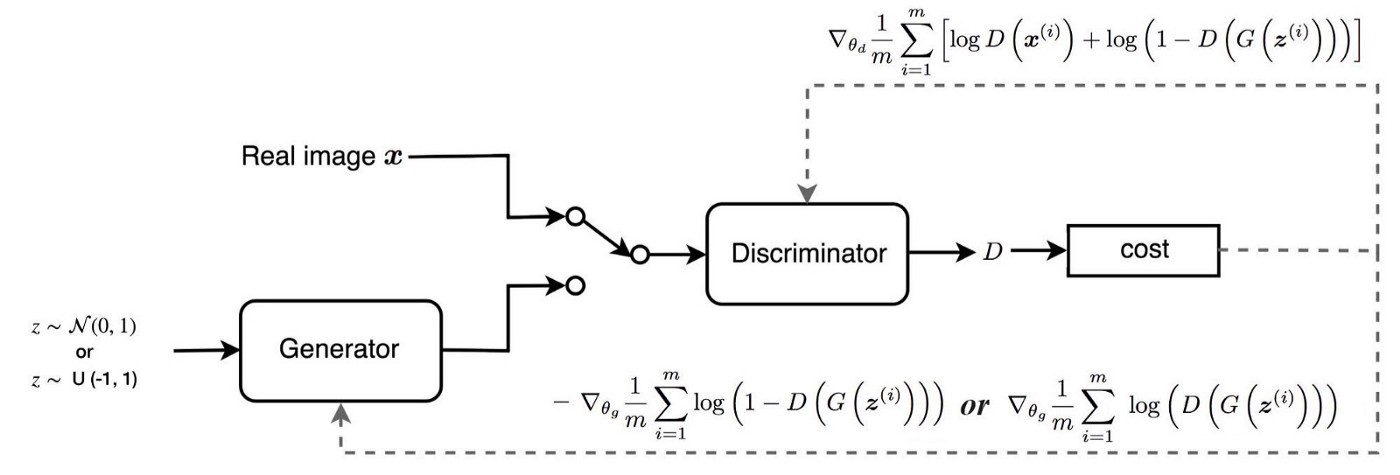}
			\caption{Generative Adversarial Network (GAN)[46]}
			\label{Generative Adversarial Network}
		\end{centering}
	\end{figure}
	
	Generative adversarial networks take up a game-theoretic approach. The network learns to generate from a training distribution. The Generator G$(z)$ takes random noise vector $z$ as input with corresponding network weights. The Discriminator $D$, on the other side, is a classifier with a containing loss function. The Discriminator has to determine if the sample is real data $x$ or generated data G$(z)$. Depending on whether the Discriminator has evaluated the input correctly or wrong, it gets an update on the Generator's parameters. The Generator and the Discriminator can perform almost the same tasks. Nonetheless, there is a difference in the math behind. The generative model describes how data is generated to learn any kind of data distribution using supervised learning. The Discriminator, however, models the decision boundary between the classes. Then it outputs the probability that $x$ comes from the data rather than from the probability distribution $p_g$.
	So both models predict the conditional probability but from different probabilities. [27] Each function is differential concerning its inputs and with respect to its parameters. [26] The overall value function V$(D$, $G)$ for the connection between $G$ and $D$ is defined as:
	
	\begin{equation}
	E\textsubscript{x}[\log(D(x))] + E\textsubscript{z}[\log(1-D(G(z)))]
	\end{equation}
	
	D($x$) is the probability, given by the Discriminator, that real data instance x is real. $E_x$ is the expected value over the real data instances. $G(z)$ is the output of the Generator from a given noise $z$.
	The input $z$ to the Generator is sampled from some simple noise, such as the uniform distribution or a spherical Gaussian distribution. They should be chosen for the following reasons [28]:
	
	\begin{enumerate}
		\item{The unit variance means that each element in the noise vector can be a different feature of the output image.}
		\item{With the Gaussian distribution, it will be easy to draw samples and even interpolate between two values and see a progressive change in the image.}
	\end{enumerate}
	
	D$(G(z))$ is the Discriminator's estimate of the probability that a fake instance is real. $E_z$ is the expected value over all random inputs to the Generator. In other words, it is the expected value over all generated fake instances $G(z)$. The Discriminator is trained to maximize the probability of choosing the real label. The Generator, however, is trained to minimize $\log(1-D(G(z)))$. The Generator $G$ cannot directly affect $\log(D(x))$ in the function, but it can minimize the loss, which is equivalent to reduce $\log(1-D(G(z)))$. During training, G$_z$ is performing not so well and is just generating some random noise, so D$(x)$ can reject samples from G($z$) with great confidence. Both optimize their weights according to their opposing loss function objective. The network improves each other until the Generator generates data that are indistinguishable from real data. [27] 
	
	\newpage
	\subsection{Discriminator loss}
	
	When the Discriminator is trained, it decides both the real data and the fake data from the Generator. It penalizes itself for declassifying a real instance as fake, or a fake instance as real. It punishes itself if a real instance is evaluated as as fake, or a fake, created by the Generator, is evaluated as real. It can achieve it by maximizing the below function. [27]
	
	\begin{equation}
	\nabla\theta\textsubscript{d}\nicefrac{1}{m}\sum[\log D(x\textsuperscript{(i)}) + \log(1 - D(G(z\textsuperscript{(i)})))]
	\end{equation}
	
	$\log(D(x))$ refers to the probability that the Generator is classifying the real image, maximizing $\log(1-D(G(z)))$ may help to label the fake image correctly.
	
	\subsection{Generator loss}
	
	The Generator takes random noise and then produces an output from that noise. The Generator loss is calculated from the Discriminator’s classification and if it successfully fools the Discriminator, it gets rewarded, otherwise, it gets penalized. [52] The following equation gets minimized to training the Generator:
	
	\begin{equation}
	\nabla\theta\textsubscript{g}\nicefrac{1}{m}\sum[\log(1 - D(G(z\textsuperscript{(i)})))]
	\end{equation}
	
	\subsection{Challenges in the training process}
	
	In the first step, the Discriminator is solid. So no adjustments to the parameters are made for the Discriminator network at this moment. The Generator get trained for a certain number of so-called training steps. The Generator is trained by using backpropagation. The training progress depends on the current state of the Discriminator, because various problems can occur in training a GAN. One of them is that the feedback of the Discriminator gets worse, as the Generator is performing better during the training. The training has to be finished in time, otherwise, the quality of the Generator and the Discriminator may decrease due to the random return values of the Discriminator. [10]
	
	\subsubsection{Training of the Discriminator}
	
	The Generator model's weights and biases are constant when creating samples for the Discriminator to get trained. The Discriminator is related to two loss functions, the Discriminator loss and Generator loss. During training of Discriminator, the Discriminator uses the Discriminator loss but ignores the Generator's loss. It attempts to classify the original data as 1 and 0 for a fake. Otherwise, it predicts a probability value. Based on the Discriminator loss, the Discriminator get penalized for misclassification. In a manner of backpropagation, the Discriminator’s weights are updated using the Discriminator loss. [55]
	
	\subsubsection{Training of the Generator}
	
	Based on the Discriminator output, the Generator loss is calculated to obtain gradients and penalize the Generator for failing in fooling the Discriminator.
	The Generator's objective is to generate fake samples that the Discriminator classifies as original and predicts a probability of 0.5 as it cannot differentiate between fake or original samples. During Generator training, the Discriminator training is stopped, and its weights remain fixed to avoid the Discriminator model from becoming too strong. If the Discriminator training is kept on during Generator training, the Generator will never converge or be able to learn the distributions. This is because of the Discriminator, which will continue to improve and then would become too good in recognizing the fake samples. [55]
	
	\subsubsection{Mode Collapse}
	
	Another common problem is the “mode collapse”. It may occurs that the Generator collapse to a setting where it always will produce the same outputs. It is a familiar failure case for a GAN. Even though the Generator might fool the Discriminator, it fails to present the complex real-world data distribution. It therefore gets stuck in a small space with a low variety. The classification performance of the Discriminator does not get affected by the low variety in the generated samples because the low variety is in the real data. The Discriminator, therefore, has no control getting the Generator to increase its variety. [10]
	
	\subsubsection{Vanishing problem}
	
	The Vanishing gradient problem is a difficulty in training artificial neural networks with gradient-based learning methods and backpropagation. In such scenarios, each of the neural network's weights receives an update proportional to the partial derivative of the error function concerning the current weight in each iteration of training. In some cases, the gradient will be little by preventing the weight from changing its value. This may stop the neural network from training and gets stuck. [50]
	
	\subsection{Implementation of a GAN}
	
	\begin{figure}[ht]
		\begin{centering}
			\includegraphics[width=13cm]{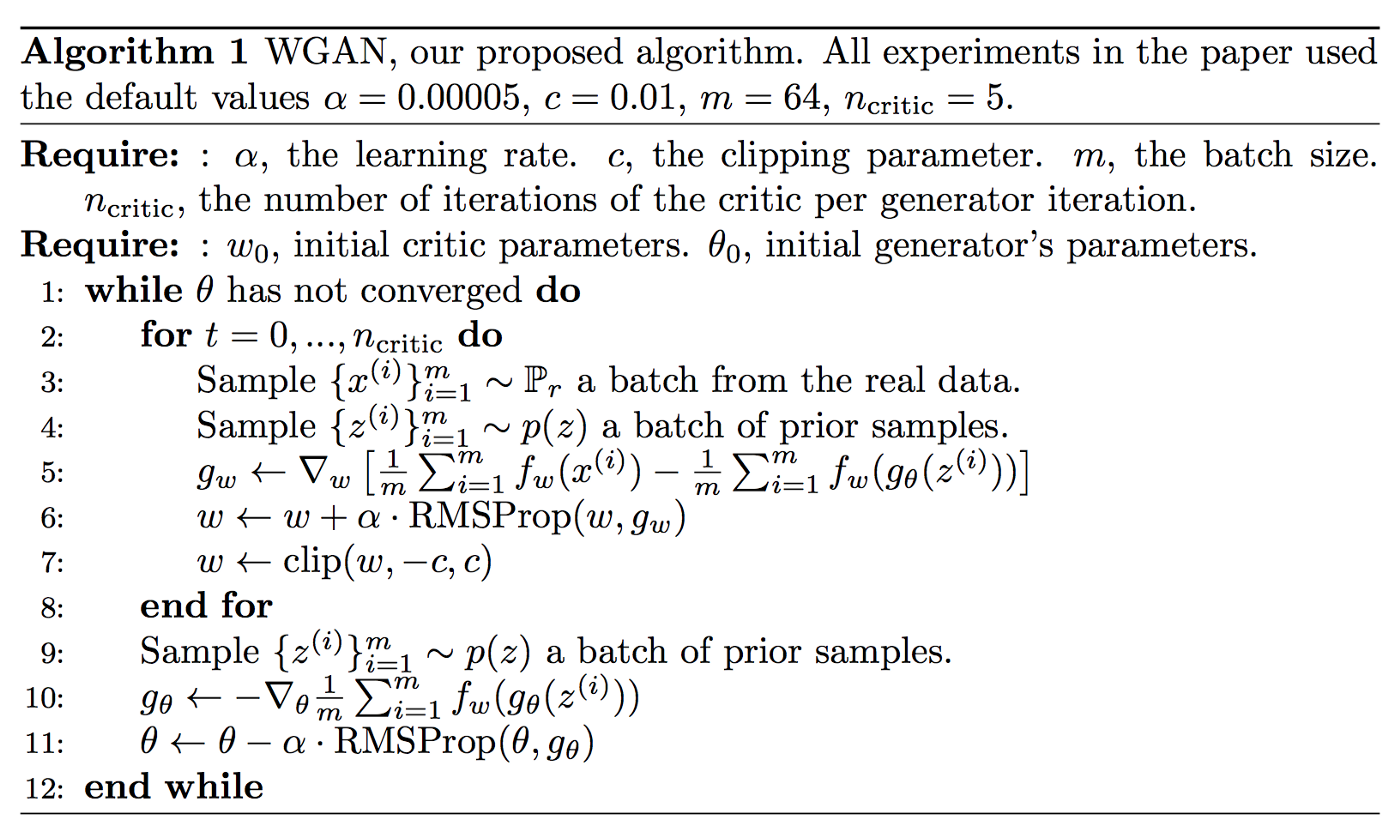}
			\caption{Implementation of GAN}
			\label{Implementation of GAN}
		\end{centering}
	\end{figure}
	
	\subsection{The Wasserstein GAN}
	\begin{figure}[ht]
		\begin{centering}
			\includegraphics[width=12cm]{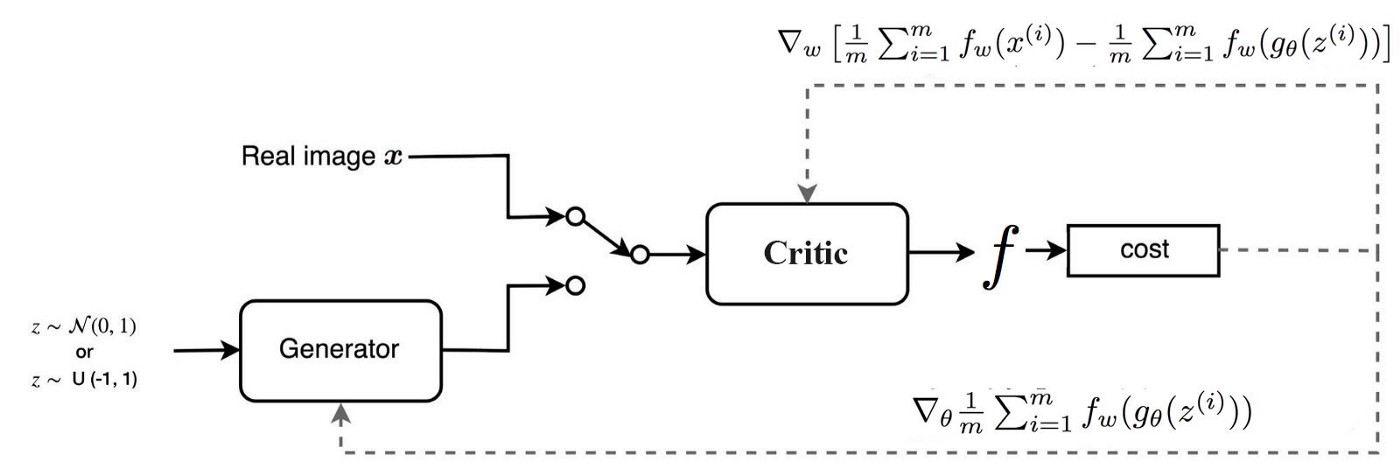}
			\caption{Model of a WGAN}
			\label{Implementation of WGAN}
		\end{centering}
	\end{figure}
	
	The main differences between a standard GAN and a WGAN:
	\begin{itemize}
		
		\item{A WGAN uses the Wasserstein loss.}
		
		\item{The WGAN is trained using labels of 1 for real and –1 for fake.}
		
		\item{There is no need for a sigmoid activation in the final layer in the WGAN critic.}
		
		\item{Train the critic multiple times for each update of the Generator.} 
		
	\end{itemize}
	
	The Wasserstein GAN was one of the first significant steps toward stabilizing GAN training. With some changes, the authors were able to show how to train GANs that have the following two properties (quoted from the paper):
	
	\begin{itemize}
		\item{A meaningful loss metric that correlates with the Generator’s convergence and sample quality}
		
		\item{Improved stability of the optimization process}
	\end{itemize}
	Specifically, the paper introduces a new loss function for both the Discriminator and the Generator. It is using the Wasserstein loss function instead of binary cross-entropy from the original GAN results in a more stable convergence of the model.
	It is defined as the shortest average distance to move the probability mass from one distribution to another. Two distributions which are in lower-dimensional manifolds without overlaps, the Wasserstein distance can provide a smooth representation of the space in-between, which is not always the case with the JS-Divergence.
	
	\begin{figure}[!ht]
		\begin{flushright}
			\begin{tabular}{c@{\hskip .7cm}c@{\hskip .8cm}c@{\hskip 1cm}r@{}}
				& \textbf{Discriminator/Critic} & \textbf{Generator} & \\
				&&&\\
				\textbf{GAN} & {\small $\displaystyle \nabla_{\theta_{d}}\frac{1}{m}\sum_{i=1}^{m}{\left [ \log D \left ( x^{\left ( i \right )} \right ) + \log \left ( 1 - D \left ( G \left ( z^{\left ( i \right )} \right ) \right ) \right ) \right]}
					$} & {\small $\displaystyle \nabla_{\theta_{g}}\frac{1}{m}\sum_{i=1}^{m}{\log\left ( D \left ( G \left ( z^{\left ( i \right )}\right )\right )\right )}$} & (\refstepcounter{equation}\theequation, \refstepcounter{equation}\theequation)\\
				&&&\\
				\textbf{WGAN} & {\small $\displaystyle \nabla_{w}\frac{1}{m}\sum_{i=1}^{m}{\left [ f \left ( x ^ {\left ( i \right )}\right ) - f \left( G \left( z^{\left ( i \right )}\right )\right )\right ]}$} & {\small $\displaystyle \nabla_{\theta}\frac{1}{m}\sum_{i=1}^{m}{f \left ( G \left ( z^{\left ( i \right )}\right )\right )}$} & (\refstepcounter{equation}\theequation, \refstepcounter{equation}\theequation)\\
			\end{tabular}
		\end{flushright}
	\end{figure}
		
	By removing the sigmoid activation from the final layer of the Discriminator, the predictions no longer fall in the range [0,1], but instead, it can now be any number in the range [$-\infty$, $\infty$]. For this reason, the Discriminator in a WGAN is usually referred to as a critic. [28] The WGAN value function is constructed using the Kantorovich-Rubstein duality to obtain:
	
	\begin{equation}
	\centering
	\underset{G}{min} \quad  
	\underset{D \in D}{max} \quad
	\underset{x\textsubscript{\textasciitilde} {\mathbb{P}\textsubscript{r}}}{\mathbb{E}} [D(x)] - \underset{\tilde{x} \textsubscript{\textasciitilde} {\mathbb{P}\textsubscript{g}}}{\mathbb{E}} [D(\tilde{x})]
	\end{equation}
	
	where $D$ are 1-Lipschitz functions and $\mathbb{P}_g$ the model distribution. Under an optimal Discriminator, minimizing the value function with respect to the Generator's parameters will minimize W($\mathbb{P}_r$, $\mathbb{P}_g$). [28] 
	The Wasserstein loss function trains the critic to convergence to ensure that the gradients for the Generator update are exact. Whereas in a standard GAN it is important to not let the Discriminator get too strong to avoid vanishing gradients. With WGANs, we can simply train the critic several times between Generator updates to make sure it is close to convergence. [56]
	
	\subsection{Lipschitz Constraint}
	
	It is unusual for allowing the critic to output any number in the range [$-\infty$, $\infty$], rather than applying e.g., a sigmoid function to restrict the output to the usual [0, 1] range. The Wasserstein loss can, for this reason, be tremendous, which is usually unsettling, due to that large numbers in neural networks should be avoided.
	The authors of the WGAN paper show that for the Wasserstein loss function, it also needs to place an additional constraint on the critic. Specifically, it is required that the critic is a 1-Lipschitz continuous function. To enforce the Lipschitz constraint on the critic, it proposes to clip the critic's weights to be in a compact space. Functions that are continuously differentiable on every point are Lipschitz continuous because of its derivation. [46]
	The critic is a function $D$ that converts an input into a prediction. A function is 1-Lipschitz if it satisfies the following inequality for any two inputs, $x\textsubscript{1}$ and $x\textsubscript{2}$:
	
	\begin{equation}
	\frac{D(x\textsubscript{1})-(x\textsubscript{2})}{\vert x\textsubscript{1} - x\textsubscript{2} \vert} \leq 1
	\end{equation}
	
	$x\textsubscript{1}$ and $x\textsubscript{2}$ are the average absolute difference between the two inputs and the critic predictions. It requires a limit on the rate at which the predictions of the critic can change. The absolute value of the gradient must be at most one everywhere. [56]
	
	\newpage
	\subsection{Wasserstein GAN with gradient penalty term}
	
	A main criticism of the WGAN is that the capacity in weight clipping of the critic is strongly diminished to learn. Even in the original WGAN paper the authors write: “Weight clipping is a bad way to enforce a Lipschitz constraint.” \newline 
	A good performance of a critic is crucial of a WGAN, because , without accurate gradients, the Generator cannot learn how to update its weights to create better samples over time. Therefore, one of the most recent extensions to the WGAN is the Wasserstein GAN Gradient Penalty (WGAN-GP) framework. The WGAN-GP Generator is defined and compiled in the same way as the Wasserstein GAN Generator. It is only the critic that needs to be changed by including the penalty gradient term.
	
	\textbf{Gradient penalty}
	
	A differential function $f$ is 1-Lipschitz if it has gradients with the norm at most 1 everywhere. So, points interpolated between the real and generated data should have a gradient norm 1 for $f$. Instead of applying weight clipping, the gradient penalty penalizes the model if the gradient norm moves away from its target norm value 1.
	
	\begin{figure}[ht]
		\begin{centering}
			{$\displaystyle
				L=\underbrace{\underset{\tilde{\bm{x}}\sim{\mathbb{P}_{g}}}{\mathbb{E}} [D(\tilde{\bm{x}})] - \underset{\bm{x}\sim{\mathbb{P}_{r}}}{\mathbb{E}} [D(\bm{x})]}_{\text{Original critic loss}}+\underbrace{\lambda\underset{\hat{\bm{x}}\sim{\mathbb{P}_{\hat{\bm{x}}}}}{\mathbb{E}}[(\|\nabla_{\hat{\bm{x}}}D(\hat{\bm{x}})\|_{2}-1)^2]}_{\text{Our gradient penalty}}
				$}
			\caption{WGAN with gradient penalty [46]}
			\label{WGAN with gradient penalty}
		\end{centering}
	\end{figure}
	
	$\tilde{x}$ is sampled from $\tilde{x}$ and x is sampled t-uniformly [0,1]. Lambda ($\lambda$) is usually set to 10. Batch normalization is avoided for the critic. Batch normalization creates correlations between samples in the same batch. By penalizing the norm of the gradient of the Discriminator with respect to its inputs instead of the existing weight clipping, it can improve classification accuracy. [28--30] 
	
	\subsubsection*{Implementation of Wasserstein with gradient penalty}
	
	\begin{figure}[ht]
		\begin{centering}
			\includegraphics[width=15cm]{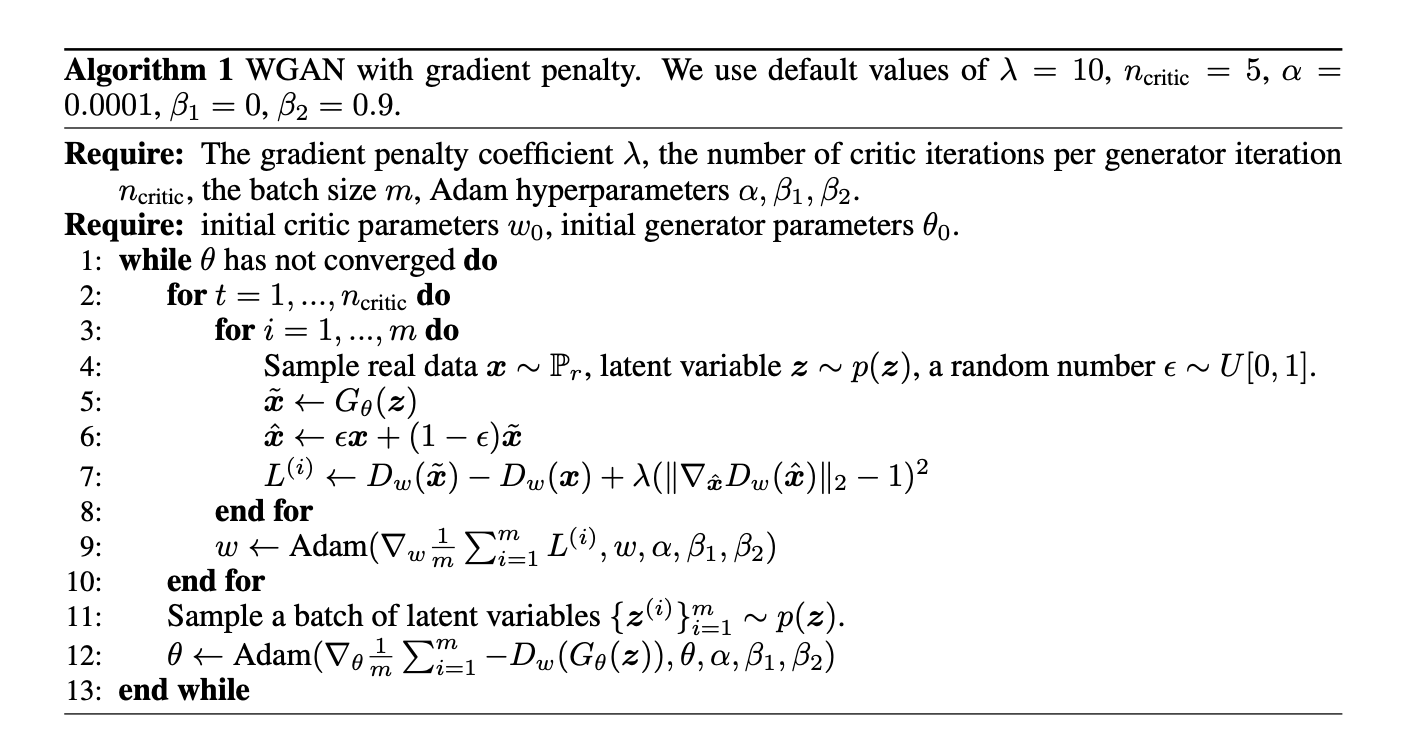}
			\caption{Implementation of WGAN-GP}
			\label{Implementation of WGAN-GP}
		\end{centering}
	\end{figure}
	
	\newpage
	\section{Theoretical evaluation methods}
	\subsection{QQ-Plot}
	
	The quantile-quantile (qq) plot is a graphical technique for determining if two data sets come from a common distribution. A qq-plot is a plot that sets the quantiles of the first data set against the quantiles of the second data set. By a quantile, it means the percent of points below a given value. If a reference line is plotted and the two groups come from a population with the same distribution, the points should fall close along this reference line. The greater the distance to this reference line, the higher the probability that the two given data sets come from populations with different distributions. There are several advantages of qq-Plot, such as:
	
	\begin{itemize}
		\item{The sample sizes do not have to be the same.}
		\item{Many distributional aspects can be simultaneously tested. As example the shifts in location and scale, changes in symmetry, and the presence of outlier can be detected in a qq-Plot.}
	\end{itemize}
	
	whereas questions can be answered as follows:
	
	\begin{itemize}
		\item{Do the two distributions from populations with a common distribution?}
		\item{Are location and scale of the two datasets familiar?}
		\item{Are the shapes of the two datasets similar?}
		\item{Are there similarities in the tails?}
	\end{itemize}
	
	When there are two data samples, the assumption of a similar distribution is justified. If so, the location and scale estimators can pool both data sets to obtain estimates of the common location and scale. When two samples are different, it can also be helpful to understand the differences. [34]
	
	\subsection{ACF}
	
	The random errors in the model are often positively correlated over time. Each random error is more likely to be similar to the previous random error that it would be, if random errors were independent of each other. The coefficient of correlation between two values in a time series is called the autocorrelation function (ACF). 
	An autocorrelation of +1 represents a perfect positive correlation, which means that an increase in one time series leads to a proportionate increase in the other time series. On the other hand, an autocorrelation of -1 represents a perfect negative correlation, so an increase in one time series results in a proportionate decrease in the different time series. 
	Even if the autocorrelation is in minus, there can still be a non-linear relationship between a time series and a lagged version of itself.
	
	\begin{figure}[ht]
		\begin{centering}
			\includegraphics[width=10.5cm]{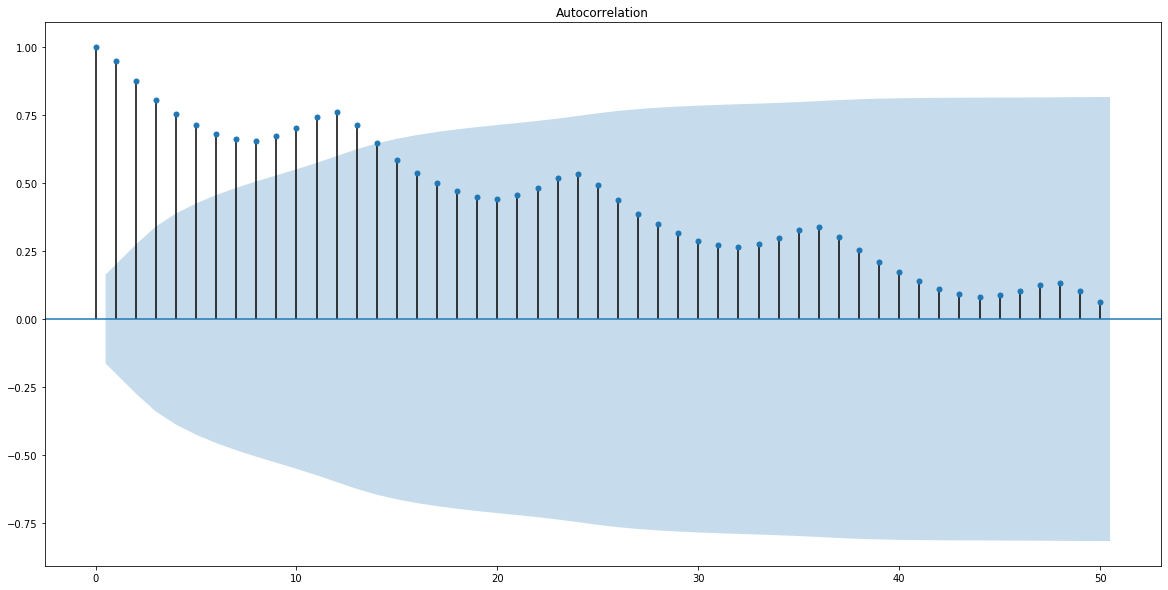}
			\caption{ACF}
			\label{ACF}
		\end{centering}
	\end{figure}
	
	\newpage
	
	\section{Descriptive Analysis}
	
	\subsection{Introduction for data and Bitcoin}
	
	\subsubsection{Bitcoin}
	
	In 2009 the Bitcoin was launched on the market and became the first publicly traded cryptocurrency. Cryptocurrencies are secure digital currencies that use encryption techniques called cryptography. Bitcoin is by far the best-known and most traded cryptocurrency with the largest market capitalization. Bitcoins are created through digital mining, in which computational puzzles are solved. The developers limited the maximum number of Bitcoins on the market to 21 million. More than 18 million of the coins have already been mined. The critical difference referring to the high risk and volatility of Bitcoin compared to traditional currencies is that Bitcoin is not supported by a central bank or backed by a government. Therefore, Bitcoin is decentralized and unregulated. Thus, monetary policy, inflation rates, and economic growth measurements, which generally impact the value of a traditional currency, do not apply to Bitcoin. In their paper, Osterrieder and Lorenz (2016) analyzed the Bitcoin exchange rates, compared them to the G10 currencies, and showed that the Bitcoin return distribution exhibits higher volatility, more substantial non-normal characteristic, and heavier tails than the G10 Currencies. 
	
	\subsubsection{Data}
	
\begin{figure}[!ht]
		\centering
		\subfloat[short for lof][Closing of Bitcoin]{
			\includegraphics[width=0.48\linewidth]{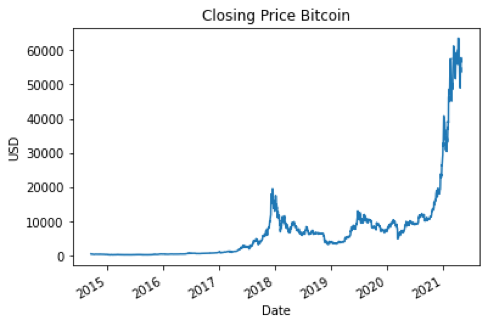}
			\label{subfig:closingbitcoin}
		}
		\subfloat[short for lof][Closings of S\&P500]{
			\includegraphics[width=0.48\linewidth]{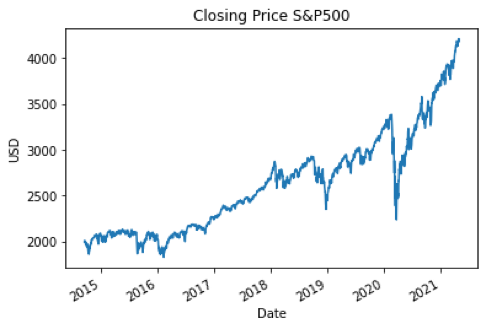}
			\label{subfig:closingsp500}
		}
		\caption[short for lof]{Closing prices of Bitcoin and S\&P500}
		\label{fig:closings}
	\end{figure}

	The data used for the Descriptive Analysis and training of the Generative Adversarial Network is obtained from yahoo finance. Although the purchase of Bitcoins is publicly available since 2009, the first time a U.S regulatory agency approved a Bitcoin financial product was in September 2014 [3]. Therefore, for this Thesis, we will use the Bitcoin publicly available data from 17.09.2014--02.05.2021 from YahooFinance.com. Further, to accomplish our study, we will run a descriptive analysis on the Bitcoin and S\&P500 to show how unique the financial time series of Bitcoin (or Cryptocurrencies in general) are. The S\&P500 is a stock index that includes the 500 largest publicly traded U.S. companies. The index is weighted by market value capitalization and is one of the world's most widely followed stock indexes. We base our analysis on the daily close prices of the assets. Note that Bitcoin trades continuously, while the S\&P500 trades on average 252 days per year. Thus, the sample size differs substantially, with 2416 data points for Bitcoin and 1666 data points for the S\&P500. This difference in trading days leads to an inevitable distortion in the following calculations. 
	Financial time series data are susceptible to random fluctuations. They are non-linear, non-stationary, and behave chaotically. Nevertheless, some observations and empirical findings are shared across various instruments, markets, and periods referred to as stylized facts. We show in Figure \ref{fig:closings} the daily close prices of Bitcoin. We see that the asset price growths generally differ. Compared to the S\&P500, Bitcoin seems to have an extremely volatile, unpredictable, and explosive growth. 
	
	The growth dynamics of the two assets generally differ. Compared to the S\&P500, Bitcoin seems to have an extremely volatile and explosive growth.
	
	\subsection{Log returns}
	
\begin{figure}[ht]
		\centering
		\subfloat[short for lof][Log return of Bitcoin]{
			\includegraphics[width=0.4\linewidth]{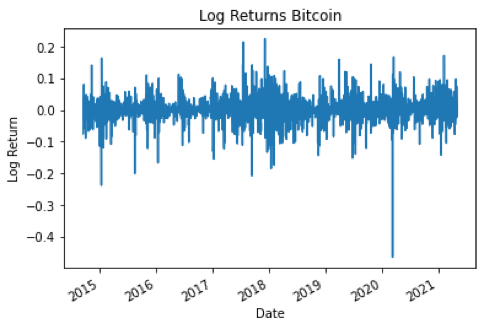}
			\label{subfig:logreturnbitcoin}
		}
		\hspace{.1\linewidth}
		\subfloat[short for lof][Log return of S\&P500]{
			\includegraphics[width=0.4\linewidth]{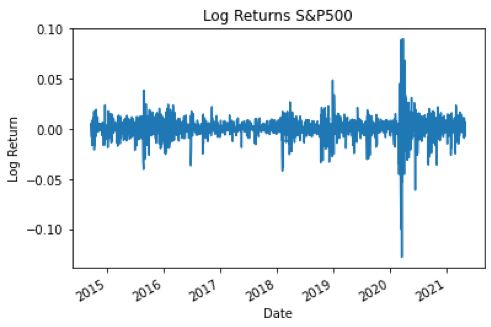}
			\label{subfig:logreturnsp500}
		}
		\caption[short for lof]{Log returns of Bitcoind and S\&P500}
		\label{fig:log returns1}
	\end{figure}

	Most methods of financial time series analysis operate with static data. A stationary time series is represented by data over time, whose statistical properties remain constant regardless of a change in the time origin. There are various ways of making time series stationary, but those ways look at the difference between values rather than the absolute values. In market data, the standard way to get stationary data is to work with log returns. They are calculated as the natural logarithm of the index today divided by the index of the day before: $\ln(\frac{V}{t-1})$. [40] Figure \ref{fig:log returns1} shows the log returns as a time series of the Bitcoin and S\&P500. We observe that the Bitcoin returns show more extreme behavior compared to the S\&P500. The extreme movements are not surprising due to the high volatility of Cryptocurrencies. 
	
\begin{figure}[ht]
		\centering
		\subfloat[short for lof][Bitcoin log returns]{
			\includegraphics[width=0.23\linewidth]{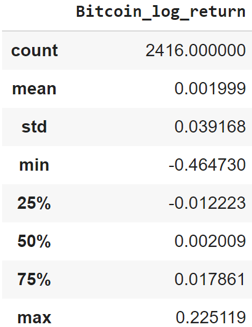}
			\label{subfig:logreturntablebitcoin}
		}
		\hspace{0.2\linewidth}
		\subfloat[short for lof][S\&P500 log returns]{
			\includegraphics[width=0.22\linewidth]{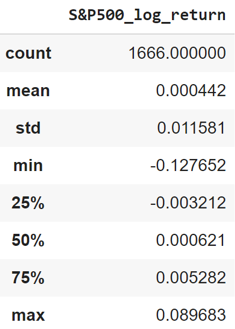}
			\label{subfig:logreturntablesp500}
		}
		\caption[short for lof]{Log return tables}
		\label{fig:log return tables}
	\end{figure}

	The following tables (Figure \ref{fig:log return tables} \subref{subfig:logreturntablebitcoin} \& \subref{subfig:logreturntablesp500}) provide a few of the essential key measures of the investigated log returns. As you can see in the table, the log returns minimum, maximum, and quantile values of Bitcoin are much more extreme than the S\&P500 values. The volatility of a financial time series is often referred to as the standard deviation (std). Comparing the standard deviations in the table shows that the Bitcoin behaved (in the observed period) over three times more volatile than the S\&P500. It is interesting to note that traditional currencies typically exhibit lower volatility than stock indexes. [39][43]
	
	\subsection{Fat tails}
	
\begin{figure}[htbp]
		\centering
		\subfloat[short for lof][Q-Q plot of log return Bitcoin]{
			\includegraphics[width=0.36\linewidth]{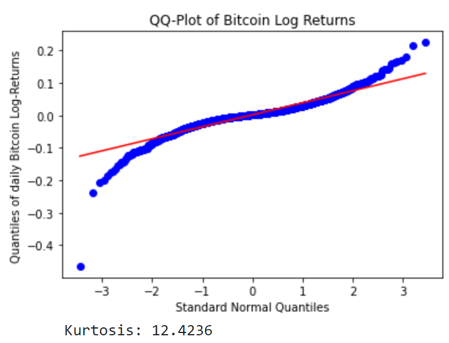}
			\label{subfig:qqplotbitcoin}
		}
		\hspace{.12\linewidth}
		\subfloat[short for lof][Q-Q plot of log return S\&P500]{
			\includegraphics[width=0.36\linewidth]{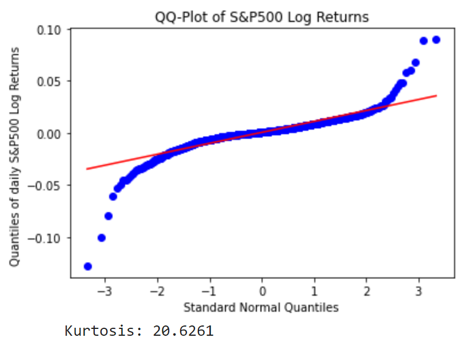}
			\label{subfig:qqplotsp500}
		}
		\caption[short for lof]{QQ-Plots of Log return of Bitcoin and S\&P500}
		\label{fig:qqplots}
	\end{figure}

	A common stylized fact of financial time series is the exhibition of fat tails. The distribution of a variable has fat tails if its outcomes are more extreme than a normally distributed variable, where the mean and variance are equal. The qq-plot is a common graphical method that is used to analyze the tails of a distribution. Figure \ref{fig:qqplots} show the qq-plots, comparing the distribution of log returns with the expected normal distribution. We see that Bitcoin and S\&P500 returns generally do not correspond to a normal distribution.

	\subsection{Volatility Clusters}
	
\begin{figure}[htbp]
		\centering
		\subfloat[short for lof][ACF of Bitcoin Log Returns]{
			\includegraphics[width=0.36\linewidth]{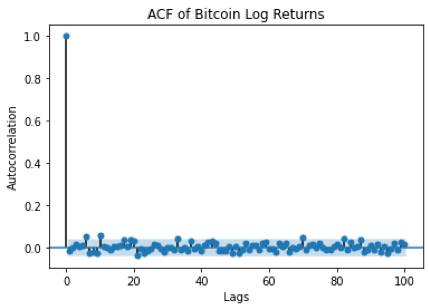}
			\label{subfig:ACF_bitcoin_new}
		}
		\hspace{.12\linewidth}
		\subfloat[short for lof][ACF of S\&P500 Log Returns]{
			\includegraphics[width=0.36\linewidth]{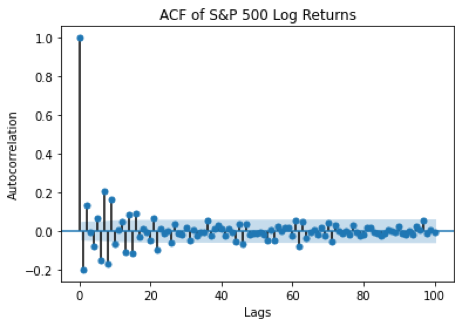}
			\label{subfig:ACF_sp500_new}
		}
		
		\subfloat[short for lof][ACF of absolute Bitcoin Log Returns]{
			\includegraphics[width=0.36\linewidth]{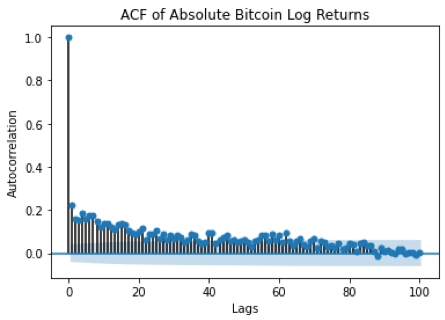}
			\label{subfig:acf_abs_bitcoin}
		}
		\hspace{.12\linewidth}
		\subfloat[short for lof][ACF of absolute S\&P500 Log Returns]{
			\includegraphics[width=0.36\linewidth]{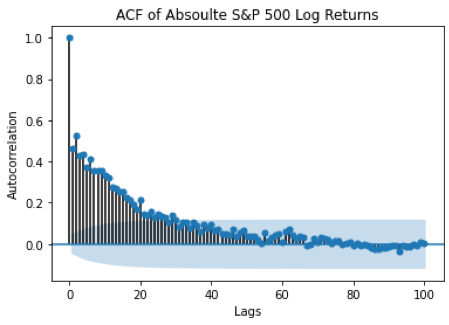}
			\label{subfig:acf_abs_sp500}
		}
		\caption[short for lof]{ACF of absolute and average Bitcoin and S\&P500 returns}
		\label{fig:acf}
	\end{figure}

	A common stylized fact is volatility clustering. It means that high price volatility events tend to cluster with time. In addition, it generally means that significant price changes follow large changes in price levels, and small changes tend to be followed by small changes. Autocorrelation plots are realistic representations of volatility clustering effects. We use the (linear) autocorrelation function (ACF) to measure how the current price values and its past values correlate.
	In Figures \ref{fig:acf} \subref{subfig:ACF_bitcoin_new} and \subref{subfig:ACF_sp500_new}, we attest that no such linear autocorrelations exist neither among Bitcoin nor across the S\&P500 data sets. This result confirms generally accepted knowledge that the price movements in financial markets are not exhibiting any significant linear correlations. If such correlation were statistically significant, the prediction of stock prices would overall deliver highly accurate results because there would be strong evidence for predictability.
	In Figures \ref{fig:acf} \subref{subfig:acf_abs_bitcoin} and \subref{subfig:acf_abs_sp500}, the dependency of the conditional variance can be captured using absolute log returns. Moreover it shows, that absolute log returns present significant correlations with a slow decaying trend. Therefore,  we conclude that log returns are not independent, principally caused by volatility clustering. [41]
	
	\newpage
	\subsection{Skewness and kurtosis}
	
\begin{figure}[ht]
		\begin{centering}
			\includegraphics[width=11.5cm]{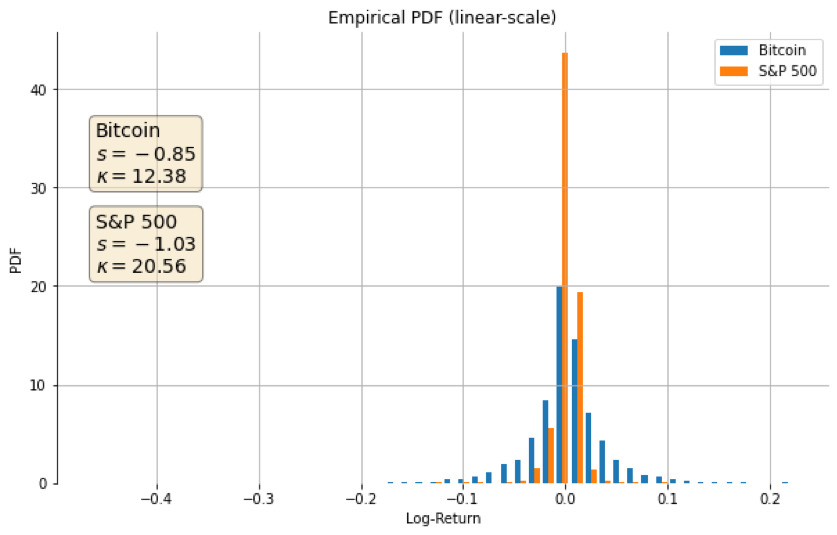}
			\caption{Empirical evaluation of Bitcoin and S\&P500}
			\label{fig:skew_kurt}
		\end{centering}
	\end{figure}

	Fundamental statistic and probability theory almost exclusively deal with the first and second central moment of a random variable, expectation and variance. For the statistical analysis of financial data, the third and fourth central moments called skewness and kurtosis are often meaningful. \newline The skewness is the third central moment and is essential for measuring another common stylized fact called the "Gain-Loss Asymmetry". It is referred to as the observation when significant drawdowns in a stock index value appear, but not equally large upward movements. The skewness measures the symmetrical gathers around its mean, and therefore are all symmetrically characterized with a skewness value of 0. Positive skewness implies that the right tail is fatter than the left, which means that positive returns tend to occur more often than significant negative returns. The kurtosis is the fourth central moment and measures the degree of peakedness of the distribution relative to its tails. The kurtosis is a measurement mix between asymmetry and tail-weight, and hence it is more informative for symmetrical distributions. The kurtosis of the normal distribution is given by the value 3. This is the reason,  why a higher kurtosis value than 3 is a signal for fat tails. \newline
	The minimal kurtosis value of a distribution is equal to 1. The probability density function of the log-returns of both investigated distributions is shown in Figure \ref{fig:skew_kurt} with their associated skewness and kurtosis values. At first glance, we can see that the log-returns of Bitcoin are more spread and the S\&P500 and has a higher peak. The negative skewness values imply that both distributions have fatter left tails than right tails, whereas the S\&P500 seems slightly more left-skewed. Additionally, we attest that both datasets have a higher kurtosis value than the normal distribution. Our analysis shows that the value of the S\&P500 has a higher kurtosis value than Bitcoin.  
	
	\newpage
	\section{Our approach to Wasserstein GAN with gradient penalty and LSTM}
	
	Our Implementation of the WGAN-GP written with TensorFlow 2 (Version 2.4.1 and Keras Version 2.4.3) is inspired by the Keras-Team who have created a Wasserstein GAN on images as described in paper [28].
	We simply had to adjust/lower the input dimensions of tensorflow/numpy vectors and Generator and Discriminator architecture to use it on time series.
	
	\subsection{Data processing and Implementation}
	
	Note that the gradient w.r.t of the input averaged samples, not the Discriminator weights, that we are penalizing. To evaluate the gradients, we must first run samples through the Generator and evaluate the loss. Then we get the gradients of the Discriminator w.r.t. the input averaged samples. The l2 therefore calculates the norm and penalty. This loss function requires the original averaged samples as input, but Keras only supports passing y$\textsubscript{true}$ and y$\textsubscript{pred}$ to loss functions.
	
	The keras-team made a partial of the function with the averaged samples argument, and used that for model training. The Keras API merge layer function to calculate the weighted average for the interpolation did not work on Tensorflow Versions above 1.15. So we had to create a custom Merge layer using the class tensorflow.keras.layers.Layer that can perform a weighted sum/merge of two different layers' outputs to get the weighted average. As for the training we made three different labels of vectors. Valid is the vector label for real samples with value 1, the fake is the label vector for generated samples, with value -1 and a dummy sample (not used) which is passed to the gradient penalty loss function.
	In particular, the Implementation has two parts. One Discriminator and one Generator. The implemented training is in two steps. The Discriminator will be trained. During the training of the Discriminator Generator layers are frozen. The Discriminator takes both real image samples and random noise seeds as input. The noise seed is run through the Generator model to get generated images.
	
\begin{figure}[ht]
		\begin{centering}
			\includegraphics[width=15cm]{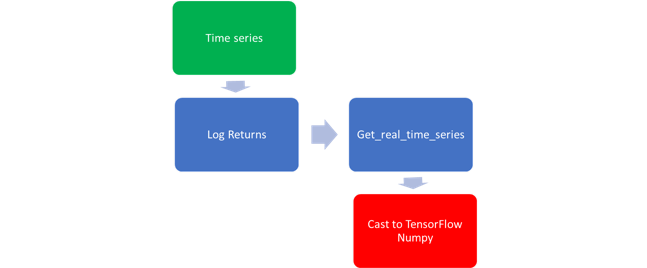}
			\caption{Data-processing}
			\label{fig:pipeline}
		\end{centering}
	\end{figure}

	Both real time series (processed in the pipeline as shown in Figure \ref{fig:pipeline}) and the generated images are then run through the Discriminator. After training of the Discriminator the Generator model is used to train the Generator layers. As such, the Implementation ensures that the Discriminator layers are not trainable. 
	The whole training is organized in two loops. In the WGAN model, the critic model must be updated more than the Generator model. A new hyperparameter, called n$\_$critic, is defined to control the number of times that the critic is updated for each update to the Generator model, and is set to 5. In Figure \ref{fig:pipeline}, we show out data processing pipeline.
	
	\newpage
	\subsection{Parameter settings}
	
	\begin{figure}[ht]
		\begin{centering}
			\includegraphics[width=5cm]{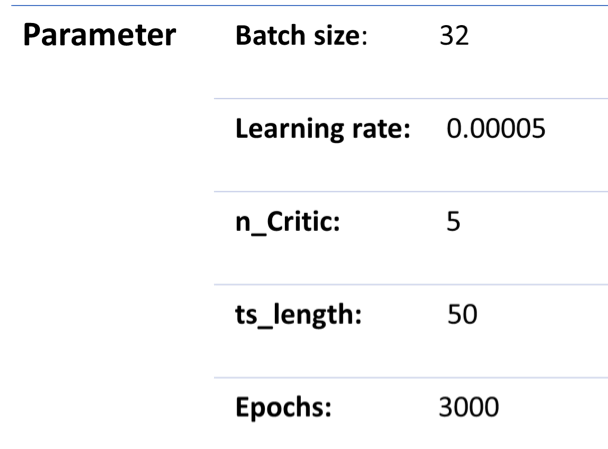}
			\caption{Parameter settings}
			\label{fig:parameter-settings}
		\end{centering}
	\end{figure}
	
\begin{figure}[ht]
		\centering
		\subfloat[short for lof][Generator architecture]{
			\includegraphics[width=10cm]{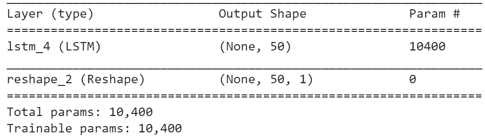}
			\label{subfig:layer1}
		}
		
		\subfloat[short for lof][Critic architecture]{
			\includegraphics[width=10cm]{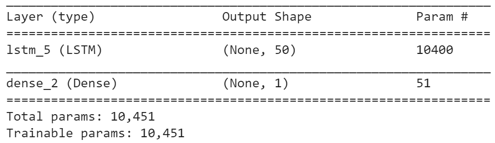}
			\label{subfig:layer2}
		}
		\caption[short for lof]{Generator and critic architecture}
		\label{fig:layers}
	\end{figure}

	In Figure \ref{fig:parameter-settings}, we demonstrate all the parameter settings that been used in our model.
	The WGAN model was trained over a runtime of 3000 epochs. The number of critic iterations per Generator iteration was set to 5, and to optimize the model we used a RMSprop optimizer with a learning rate of 0.00005. The random noise vector from which the Generator produces data has a length of 25 and is generated from a normal distribution. The generated time series has a size of 50, while the real and fake batches, which are fed to the Discriminator as inputs, consist of 32-time series each. \newline 
	The architectures of the Critic and Generator are structured as shown in Figure \ref{fig:layers}. The Generator has one LSTM block, and the output is reshaped to match the dimension of a real sample. The Discriminator has one hidden layer consisting LSTM and an output layer of dimension one. The output of the Discriminator is just without the sigmoid function and outputs a scalar score rather than a probability. This score can be interpreted as how real the input images are.

	\subsection{Architecture}
	
	For the Discriminator and also for the Generator we are using vanilla LSTM. A Vanilla LSTM is an LSTM model with a single hidden layer of LSTM units and an output layer used to make a prediction. LSTMs can be used to model univariate time series forecasting problems. These are issues comprised of a single series of observations, and a model is required to learn from the series of past observations to predict the next values in the sequence. 
	LSTM adds the explicit handling of sequence between observations when learning a mapping function from inputs to outputs, not offered by MLPs or CNNs. They are a type of neural network that adds native support for input data comprised of sequences of observations. The shape of the input in each sample is specified by the input shape argument of LSTM layer. We almost always have multiple samples. Therefore, the model will expect the input component of training data to have the dimensions or shape: 
	\textbf{[samples, timesteps, features]}
	
	The basic LSTM neural network cell is something like $h\textsubscript{t}=f(W\textsubscript{h}h\textsubscript{t-1}+W\textsubscript{x}x\textsubscript{t})$. [70] So, it takes previous hidden state $h\textsubscript{t-1}$ and current input $x\textsubscript{t}$ to produce hidden state ht and in each time step, we use the same weights. The specified number of time steps defines the number of input variables used to predict the next-time step. We are using 50-time steps. For daily data, this is about a two-month time series sequence. \newline
	The selection of 50-time steps was inspired by the 50 day "MA" (moving average) in stock trading. The 50-day moving average indicator is one of the most important and commonly used tools in stock trading. The 50-day moving average is one of the leading indicators. It is, therefore, the first line of major moving average support in an uptrend or the first line of major moving average resistance in a downtrend. That is why we use 50-time steps to train our LSTM. [69] Since this is a univariate time series prediction, we have the time series itself. That's why the feature is set to one. It is also possible to consider as additional feature the autocorrelation of the time series itself.
	The LSTM network in both Generator ($G$) and Discriminator ($D$) has depth 1; each LSTM cell has 50 units. The output from each LSTM cell in $D$ are fed into a fully connected layer with weights shared across time steps, and one output per cell is then averaged to the final decision for the sequence. \newline
	Using unidirectional LSTM only preserves the information of the past because the only inputs it has seen are from the past.
	Using bidirectional will run your inputs in two ways, one from past to future and one from future to past and what differs this approach from unidirectional is that in the LSTM that runs backward you preserve information from the future and using the two hidden states combined, you can preserve information from both, past and future, in any point in time. Since there is no economic rational to use future information in the past and the possibility that the model starts to learn noise from the future, we did not choose bidirectional LSTM.
	Generator and Discriminator have both the same LSTM architecture since they need to learn the same sequence pattern in the time series.
	
	\subsection{Experimental Results}
	
\begin{figure}[ht]
		\centering
		\subfloat[short for lof][Epoch 500]{
			\includegraphics[width=0.3\linewidth]{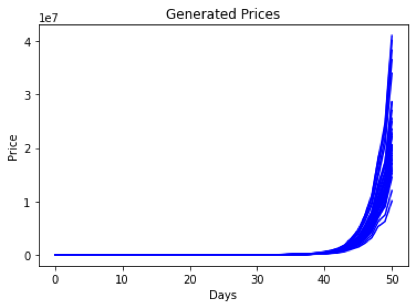}
			\label{subfig:Epoch500samples}
		}
		\subfloat[short for lof][Epoch 1000]{
			\includegraphics[width=0.3\linewidth]{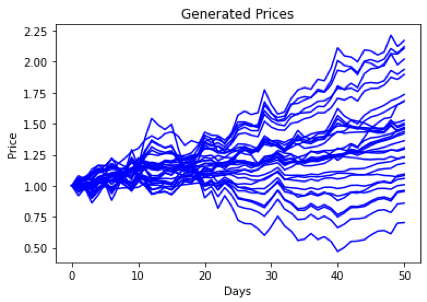}
			\label{subfig:Epoch1000samples}
		}
		\subfloat[short for lof][Epoch 1500]{
			\includegraphics[width=0.3\linewidth]{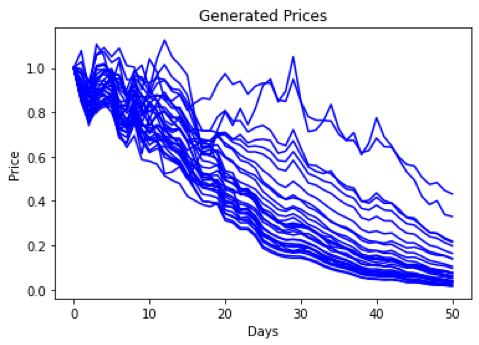}
			\label{subfig:Epoch1500samples}
		}
		
		\subfloat[short for lof][Epoch 2000]{
			\includegraphics[width=0.3\linewidth]{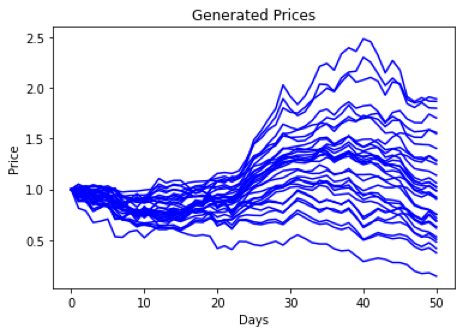}
			\label{subfig:Epoch2000samples}
		}
		\subfloat[short for lof][Epoch 2500]{
			\includegraphics[width=0.3\linewidth]{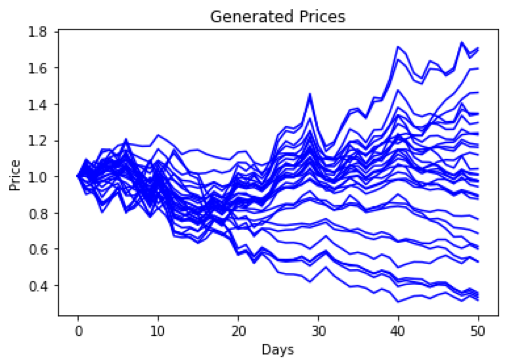}
			\label{subfig:Epoch2500samples}
		}
		\subfloat[short for lof][Epoch 3000]{
			\includegraphics[width=0.3\linewidth]{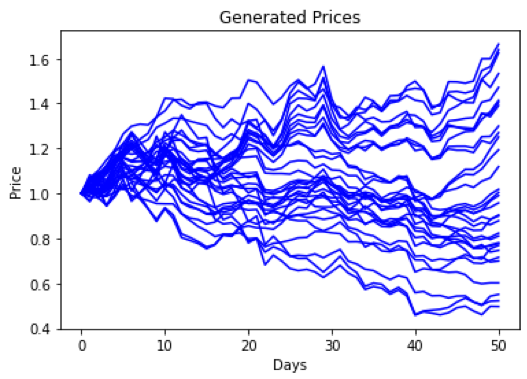}
			\label{subfig:Epoch3000samples}
		}
		\caption[short for lof]{Generated prices all 500 time steps until epoch 3000}
		\label{fig:epochs-samples}
	\end{figure}

	Figure \ref{fig:epochs-samples} are the generated samples that we collected every 500 epochs after training the model during the 3000 epochs. The results show that the model can generate specific price curves for various elements of the multitude of synthetic financial time series. Figure \ref{fig:epochs-samples} \subref{subfig:Epoch3000samples} shows all the generated artificial time series of epoch 3000 (in blue) and the real historical time series of the Bitcoin (in red). The real-time series is shown with an illustrative goal so we can make a comparison between the synthetic output and the actual data. There is not much visual difference in price movements between the artificial and real prices data sets. The synthetic data we obtain seems to follow generally the same trends as the real Bitcoin prices. This result shows that the GAN model we adapted for financial data did work. Therefore it is possible to generate synthetic representative financial data sets.
	
	\newpage
	\subsection{Distribution Comparison}
	
\begin{figure}[ht]
		\centering
		\subfloat[short for lof][Epoch 500]{
			\includegraphics[width=0.3\linewidth]{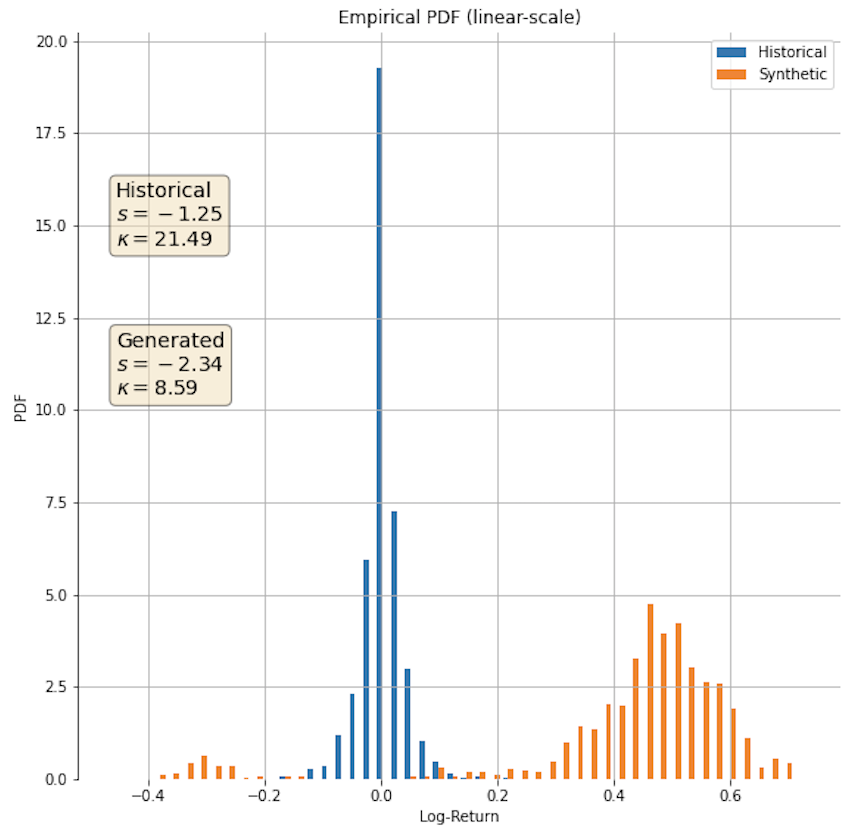}
			\label{subfig:Epoch500dist}
		}
		\subfloat[short for lof][Epoch 1000]{
			\includegraphics[width=0.3\linewidth]{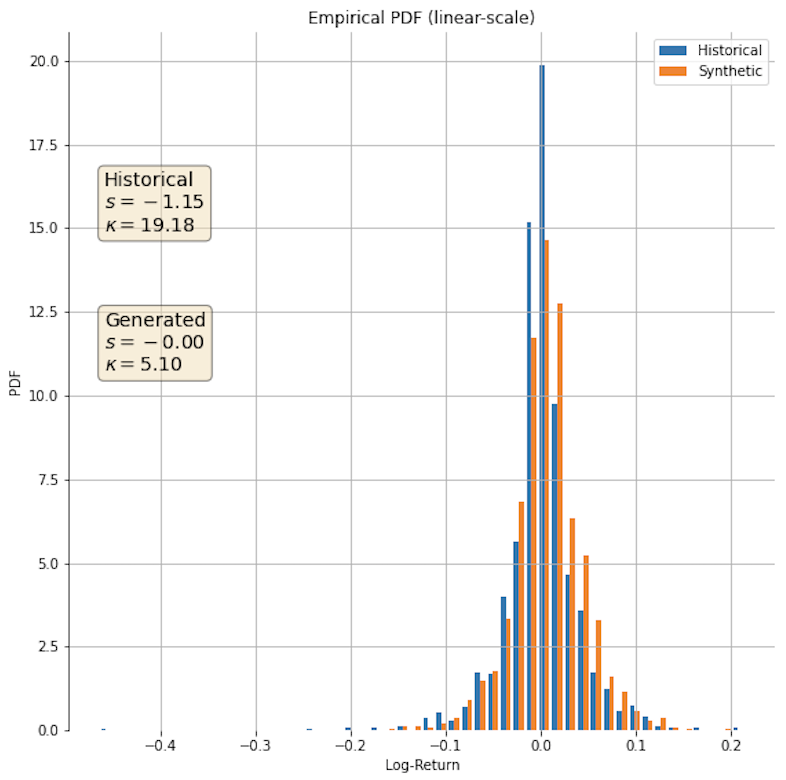}
			\label{subfig:Epoch1000dist}
		}
		\subfloat[short for lof][Epoch 1500]{
			\includegraphics[width=0.3\linewidth]{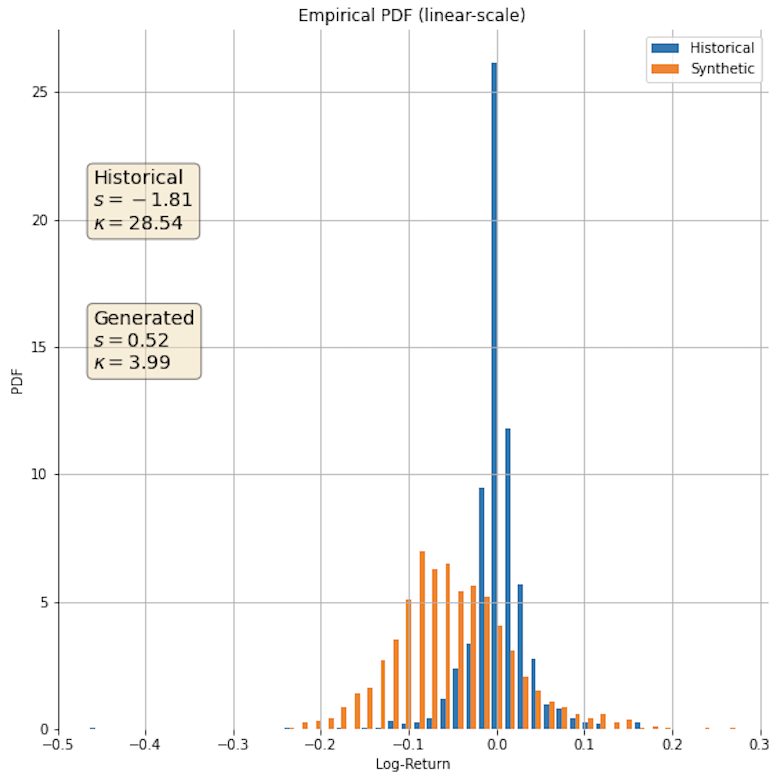}
			\label{subfig:Epoch1500dist}
		}
		
		\subfloat[short for lof][Epoch 2000]{
			\includegraphics[width=0.3\linewidth]{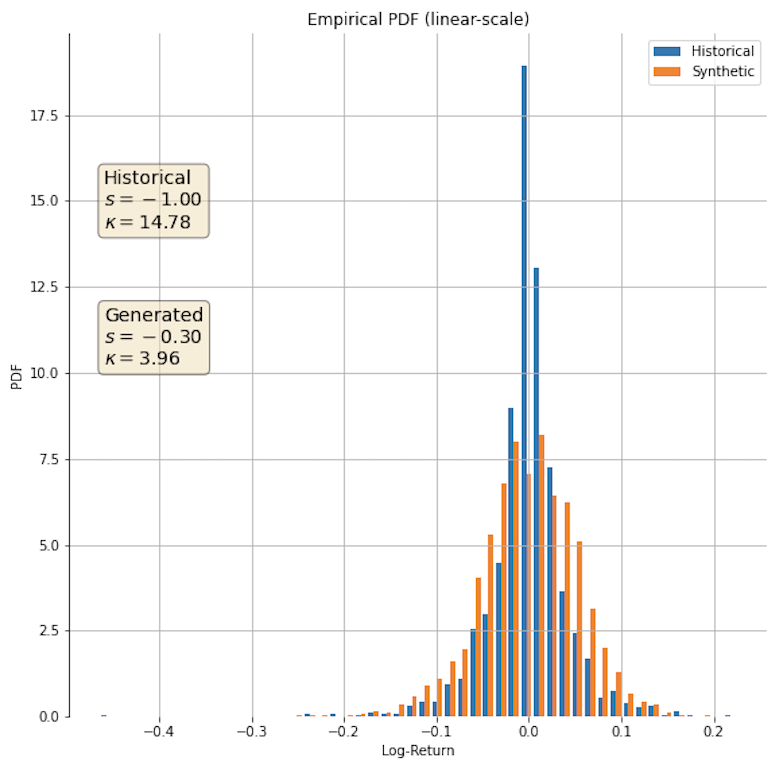}
			\label{subfig:Epoch2000dist}
		}
		\subfloat[short for lof][Epoch 2500]{
			\includegraphics[width=0.3\linewidth]{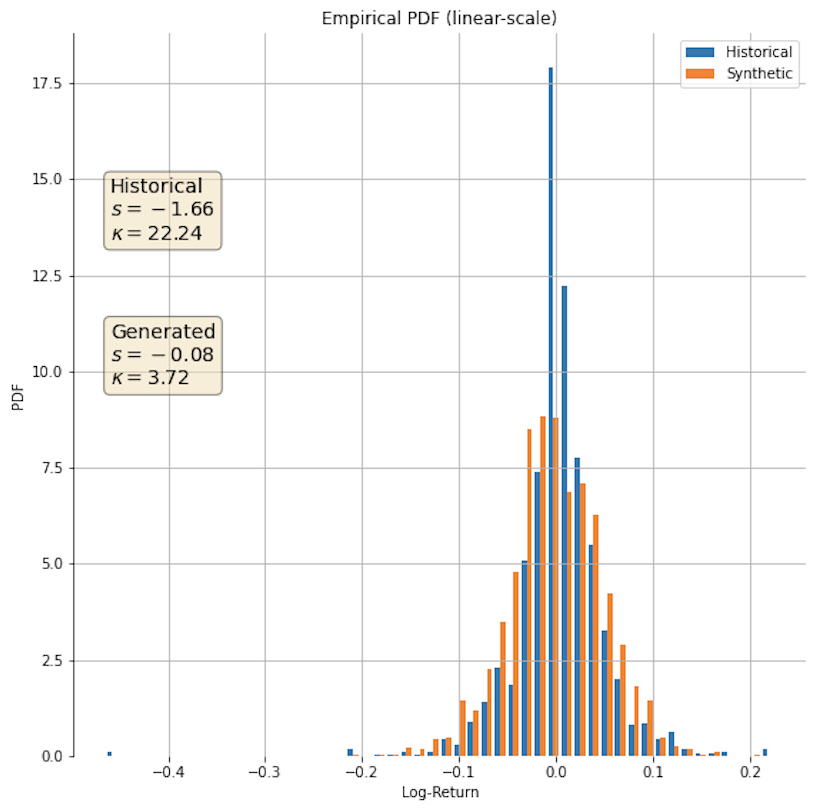}
			\label{subfig:Epoch2500dist}
		}
		\subfloat[short for lof][Epoch 3000]{
			\includegraphics[width=0.3\linewidth]{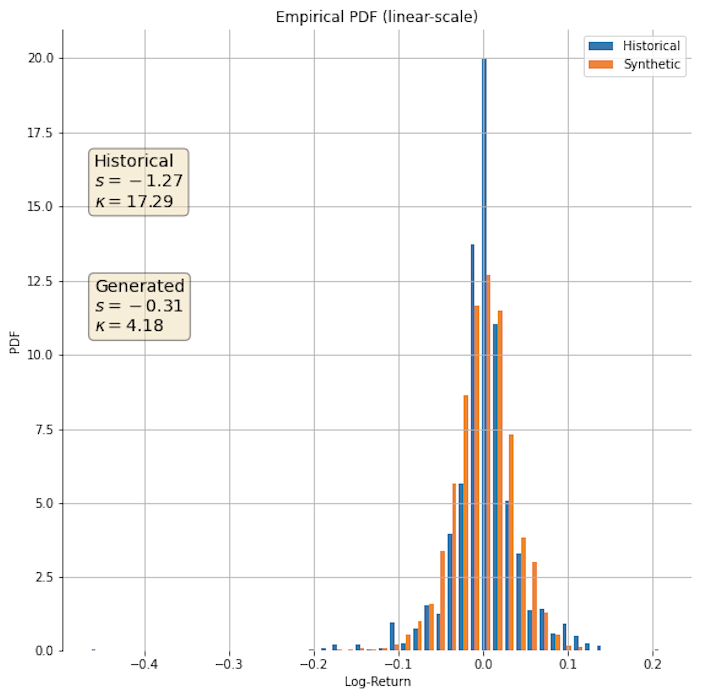}
			\label{subfig:Epoch3000dist}
		}
		\caption[short for lof]{Empirical and historical distributions all 500 time steps until epoch 3000}
		\label{fig:epochs-dist}
	\end{figure}

	Now we compare the distribution of the historical samples with the generated time series samples. Figure \ref{fig:epochs-dist} show the probability density of the log returns from the synthetic and real data passed to the Discriminator and visualized every 500 epochs. To make it more accessible to interpret the results, we represented all the log returns of one batch in a histogram (synthetic times series in red and actual Bitcoin prices in blue). Typically, the kurtosis is similar to the variance, and it measures how a distribution is spread with a focus on the tails. The skewness value measures how symmetrical distribution is gathering around its mean. The negative skewness values obtained in Epoch 3000 imply that both financial time series, synthetic and real Bitcoin log returns, are left-skewed. In Epoch 3000, the skewness value of the historical Bitcoin data, in absolute value, is considerably higher than the synthetical times series log return distribution. The result suggests that typically, historical Bitcoin incurs proportional smaller gains and higher losses than the synthetic financial time series. The kurtosis of the historical Bitcoin data presents as well a higher value than the synthetical time series. The results imply that the historical Bitcoin log returns present more substantial non-normal characteristics than synthetic time series.
	
\begin{figure}[htbp]
		\centering
		\subfloat[short for lof][Epoch 500]{
			\includegraphics[width=0.3\linewidth]{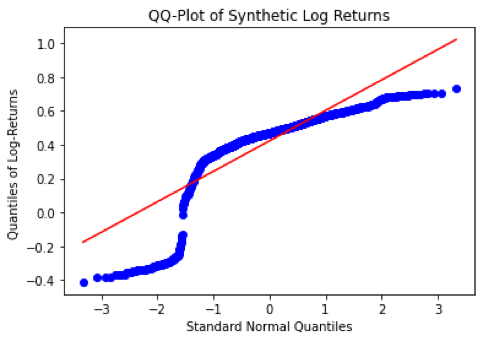}
			\label{subfig:Epoch500}
		}
		\subfloat[short for lof][Epoch 1000]{
			\includegraphics[width=0.3\linewidth]{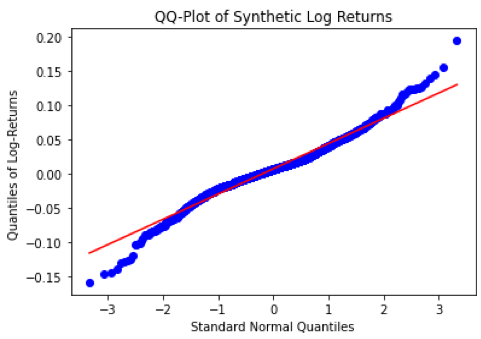}
			\label{subfig:Epoch1000}
		}
		\subfloat[short for lof][Epoch 1500]{
			\includegraphics[width=0.3\linewidth]{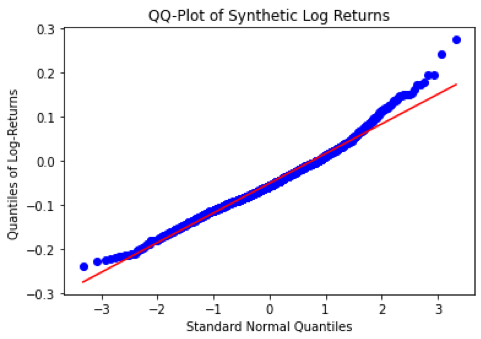}
			\label{subfig:Epoch1500}
		}
		
		\subfloat[short for lof][Epoch 2000]{
			\includegraphics[width=0.3\linewidth]{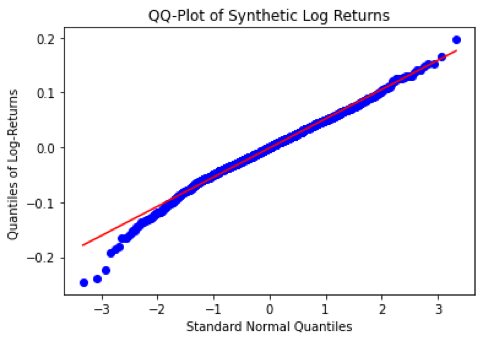}
			\label{subfig:Epoch2000}
		}
		\subfloat[short for lof][Epoch 2500]{
			\includegraphics[width=0.3\linewidth]{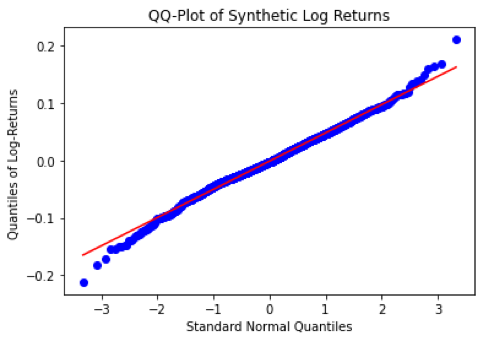}
			\label{subfig:Epoch2500}
		}
		\subfloat[short for lof][Epoch 3000]{
			\includegraphics[width=0.3\linewidth]{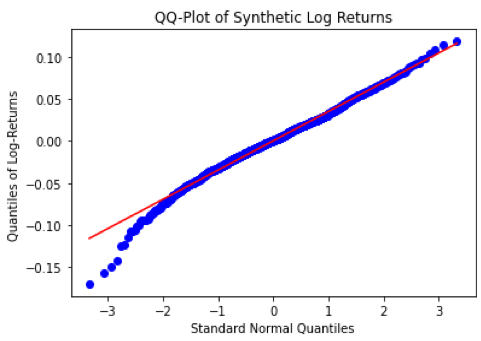}
			\label{subfig:Epoch3000}
		}
		\caption[short for lof]{QQ-Plots of Synthetic Log returns all 500 time steps until epoch 3000}
		\label{fig:epochs}
	\end{figure}

	\newpage
	As mentioned in the descriptive analysis, the historical Bitcoin log returns have fat tails as most of the stock returns. In Figure \ref{fig:epochs}, each of the quantiles of the historical and synthetic log returns of a whole epoch is plotted against the quantiles of the normal distribution. On the one hand, we observe that historical data has fatter tails compared to the tails of the artificial data illustrated in the qq-Plot of epoch 3000. The result suggests that historical Bitcoin returns are more volatile, incurring a more extreme outcome than the synthetic data. Therefore, we can attest that the WGAN we adapted to the financial times series is insufficient to generate reliable, robust data that would generally represent the underlying dataset. On the other hand, the plots show that the left tail of the generated distribution is heavier than the right tail, confirming our result that the distribution of the artificial data is left-skewed. For this reason we could say that the synthetic data is approaching the real Bitcoin log return characteristics with minimal gains and significant losses.
	
	\newpage
	\subsection{Autocorrelation (ACF)}
\begin{figure}[htbp]
		\centering
		\subfloat[short for lof][Synthetic log-returns]{
			\includegraphics[width=0.45\linewidth]{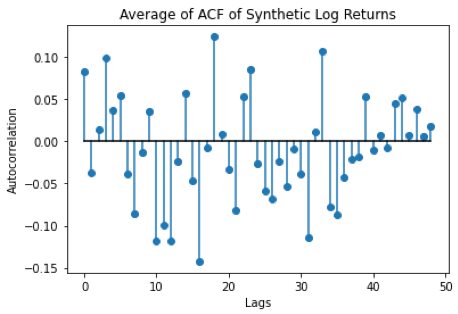}
			\label{subfig:synthetic}
		}
		\subfloat[short for lof][Real log-returns]{
			\includegraphics[width=0.45\linewidth]{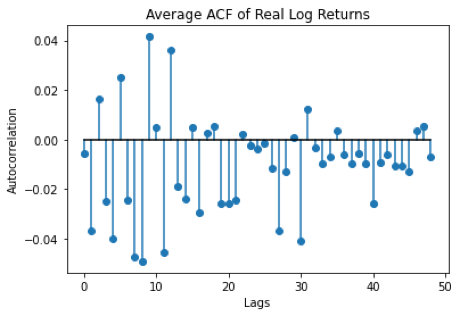}
			\label{subfig:real}
		}
		
		\subfloat[short for lof][Synthetic absolute log-returns]{
			\includegraphics[width=0.45\linewidth]{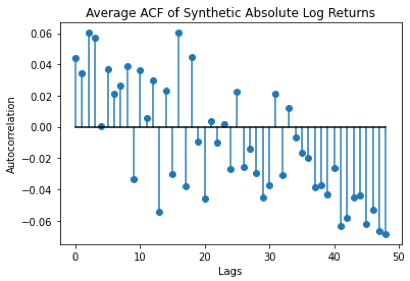}
			\label{subfig:synthetic-absolute}
		}
		\subfloat[short for lof][Real absolute log-returns]{
			\includegraphics[width=0.45\linewidth]{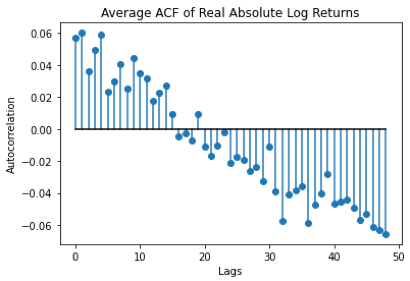}
			\label{subfig:real-absolute}
		}
		\caption[short for lof]{Average and absolute of ACF of synthetic Log returns}
		\label{fig:log returns2}
	\end{figure}

	Comparing the ACF of the log returns, we conclude that the synthetic data differs from the historical data, showing on average a higher degree of autocorrelation (see Fig. \ref{fig:log returns2} \subref{subfig:synthetic} \& \subref{subfig:real}). In Figure \ref{fig:log returns2} \subref{subfig:synthetic-absolute}, the average autocorrelations of the absolute historical log-returns show a positive and slowly decaying trend. Figure \ref{fig:log returns2} \subref{subfig:real-absolute} shows that the artificial data has a similar tendency as the real dataset but a much more fluctuating behavior. A slow decay in the ACF is an indication that a series is non-stationary because past values heavily influence the values. Since LSTM has a memory, this could explain the distinction between synthetic log-returns and real log returns. The autocorrelation could be controlled by using the length of the time series.
	
	\newpage
	\subsection{Loss function}
	
\begin{figure}[ht]
		\begin{centering}\includegraphics[width=10cm]{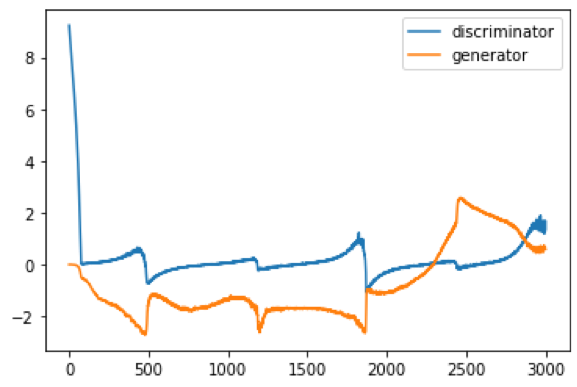}
			\caption{Loss of Generator and Critic}
			\label{fig:loss}
		\end{centering}
	\end{figure}

	In Figure \ref{fig:loss}, we plot the Loss of the Generator and Discriminator (Critic) from the adapted GAN model to the financial times series. The advantage of applied weight clipping in WGANs is that, in theory, the loss correlates with the sample quality and converges to a minimum [28]. However, these results are only proved for images, and there is no evidence of validity for the financial times series. 
	
	As illustrated in Figure \ref{fig:loss}, both the Generator loss and Discriminator loss converge and show a stable behavior while oscillating around 0. It is difficult to establish the presence of oscillation. In the absence of concrete proof, we assume the model oscillates around a minimum rather than converging to a minimum (as in theory) because the model can overrate the outliers of real time series. Besides detecting the outliers, the model strives to learn the outliers, making it challenging for the model to converge to an absolute minimum. That may be the reason why it oscillates around the absolute minimum. The loss of both gradually increases or decreases between the batches, but there is no erratic fluctuation, proving that the implemented Wasserstein metric has served its purpose. We think the stable behavior of the loss shows that the adapted WGAN model on financial time series can cope with the historical data set. The research on WGAN-GP models for generating financial time-series and evaluating their results is at an early stage. Elaborating the perfect hyperparameters of the model and setting a sufficient epoch size might lead to a decrease of oscillation or even a convergence to an absolute minimum.

	\section{Conclusion}
	
	This paper covered the structure and outline of a GAN and LSTMs in detail, summarizing the most relevant literature. We have trained our recurrent WGAN-GP model for generating real-valued sequential data. The implementation of the Wasserstein metric kept a stable loss of both the generator and discriminator during the whole training. The loss converged to a minimum with a cycle oscillation. It is difficult to find the cause of the oscillation. LSTM models are at an early stage and the model parameter settings. Finding the optimal parameters to fine-tune the model's performance is challenging and a time-consuming task that provides a basis for further investigation. However, the evaluation has shown that a visual distinction between the generated and real-time series is nearly impossible. In contrast, the statistical properties slightly differ from the generated data by more extreme outcomes. That is why we think it might be challenging for the model to catch up and replicate outliers.
	
	How close a GAN can get to the stock prices and reflect their properties is still hard to demonstrate or proof, due to its high volatility and unexpected happenings on the market. Nevertheless, promising arguments and approaches are speaking in favor of further experiments in this area. GANs seem to be an excellent methodology to capture the dynamics of financial assets and forecast future movements. Applying them to derivatives pricing, Portfolio Hedging, and Risk Management may become a crucial tool in the future since we live in Big data era.

	\newpage
	\bibliographystyle{unsrt}

\end{document}